\documentclass[journal,twoside,web]{ieeecolor}
\usepackage{generic}
\usepackage{cite}
\usepackage{amsmath,amssymb,amsfonts}
\usepackage{graphicx}

\usepackage{algorithm}
\usepackage{hyperref}
\hypersetup{hidelinks=true}
\usepackage{textcomp}
\usepackage{booktabs}
\usepackage{hyperref} 
\usepackage{rotating}
\usepackage{multirow}
\usepackage{xcolor}
\usepackage{algpseudocode}
\usepackage{xspace}

\newcommand{\algname}{BrgSA\xspace}
\def\BibTeX{{\rm B\kern-.05em{\sc i\kern-.025em b}\kern-.08em
    T\kern-.1667em\lower.7ex\hbox{E}\kern-.125emX}}
\begin{document}
\title{Bridged Semantic Alignment for Zero-shot 3D Medical Image Diagnosis}
\author{Haoran Lai, Zihang Jiang, Qingsong Yao, Rongsheng Wang, Zhiyang He, Xiaodong Tao, Weifu Lv, Wei Wei, and Shaohua Kevin Zhou,~\IEEEmembership{Fellow, IEEE}%
\thanks{Manuscript submitted on May 29, 2025; revised on September 9, 2025, and October 28, 2025; accepted on October 30, 2025. This work was supported in part by the Jiangsu Province Science Foundation for Youths under Grant BK20240464, in part by the Natural Science Foundation of China under Grant 62271465, and in part by the Suzhou Basic Research Program under Grant SYG202338. (Corresponding authors: Zihang Jiang; Wei Wei; Shaohua Kevin Zhou.)}
\thanks{Haoran Lai, Zihang Jiang, Rongsheng Wang, and Shaohua Kevin Zhou are with the School of Biomedical Engineering, Division of Life Sciences and Medicine, University of Science and Technology of China, Hefei, Anhui, 230026, China. They are also with the Suzhou Institute for Advanced Research, University of Science and Technology of China, Suzhou, Jiangsu, 215123, China (e-mail: haoranlai@mail.ustc.edu.cn; jzh0103@ustc.edu.cn; skevinzhou@ustc.edu.cn).}
\thanks{Shaohua Kevin Zhou is also with the Center for Medical Imaging, Robotics, Analytic Computing \& Learning (MIRACLE), Suzhou Institute for Advanced Research, USTC, Suzhou, Jiangsu, 215123, China; the Jiangsu Provincial Key Laboratory of Multimodal Digital Twin Technology, Suzhou Jiangsu, 215123, China; and the State Key Laboratory of Precision and Intelligent Chemistry, USTC, Hefei, Anhui, 230026, China.}
\thanks{Qingsong Yao is with Stanford University, Palo Alto, CA 94305, USA.}
\thanks{Zhiyang He and Xiaodong Tao are with the Medical Business Department, iFlytek Co. Ltd., Hefei, Anhui, 230088, China.}
\thanks{Weifu Lv and Wei Wei are with the Department of Radiology, The First Affiliated Hospital of USTC, Division of Life Sciences and Medicine, USTC, Hefei, Anhui, 230001, China. (e-mail: weiweill@ustc.edu.cn).}
\thanks{\textcolor{black}{The code, model, and detailed labeling resources are publicly available at \url{https://github.com/laihaoran/BrgSA}.}} 
}
\maketitle

\begin{abstract}
3D medical images such as computed tomography are widely used in clinical practice, offering a great potential for automatic diagnosis. Supervised learning-based approaches have achieved significant progress but rely heavily on extensive manual annotations, limited by the availability of training data and the diversity of abnormality types. Vision-language alignment (VLA) offers a promising alternative by enabling zero-shot learning without additional annotations. However, we empirically discover that the visual and textural embeddings after alignment endeavors from existing VLA methods form two well-separated clusters, presenting a wide gap to be bridged. To bridge this gap, we propose a Bridged Semantic Alignment (BrgSA) framework. First, we utilize a large language model to perform semantic summarization of reports, extracting high-level semantic information. Second, we design a Cross-Modal Knowledge Interaction module that leverages a cross-modal knowledge bank as a semantic bridge, facilitating interaction between the two modalities, narrowing the gap, and improving their alignment. To comprehensively evaluate our method, we construct a benchmark dataset that includes 15 underrepresented abnormalities as well as utilize two existing benchmark datasets. Experimental results demonstrate that \algname achieves state-of-the-art performances on both public benchmark datasets and our custom-labeled dataset, with significant improvements in zero-shot diagnosis of underrepresented abnormalities.
\end{abstract}


\section{Introduction}
\label{sec:introduction}

\begin{figure*}[t]
  \centering
    \includegraphics[width=0.95\linewidth]{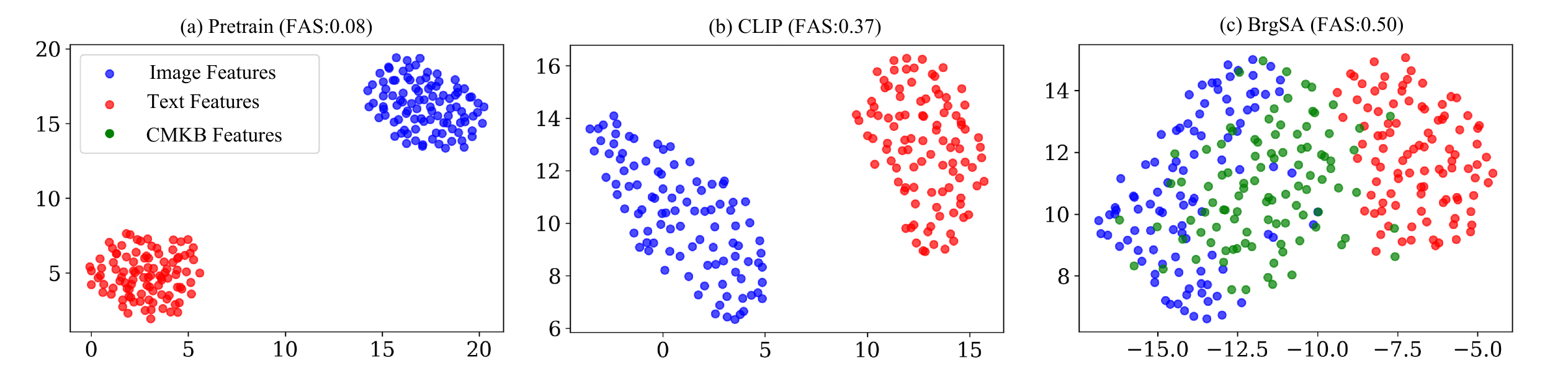}
    \caption{UMAP visualization of features. Cosine similarity is used to evaluate the alignment between image and text features. The text features are generated using generic descriptive texts to ensure that all images can be matched to all texts. (a) Features generated using pretrained weights without vision-language alignment, where image (in blue) and text features (in red) remain unaligned. \textcolor{black}{The corresponding pre-training protocols are detailed in ``\hyperref[sec:ID]{Implementation Details}''.} (b) Features after vision-language alignment using 3D CLIP, showing improved alignment but with noticeable modality gaps. (c) Features after vision-language alignment using \algname framework, where the CMKB features (in green) serve as a bridge to reduce the modality gap and further enhance feature alignment. \textcolor{black}{CMKB denotes the Cross-Modal Knowledge Bank, whereas BrgSA abbreviates Bridged Semantic Alignment.}} 
    \label{fig:motivation}
    \vspace{-4mm}
\end{figure*}

\IEEEPARstart{C}{omputed} Tomography (CT) is a widely used non-invasive diagnostic tool in clinical practice, playing a crucial role in detecting various abnormalities~\cite{ginat2014advances}. With the rapid development of artificial intelligence, significant progress has been made in leveraging CT images for automated abnormality diagnosis~\cite{svoboda2020artificial}. Supervised learning methods have achieved impressive performance in tasks such as disease classification and detection~\cite{wang2024data,mei2025clinical}. However, supervised learning methods heavily rely on large-scale, manually annotated datasets, which is time-consuming and labor-intensive~\cite{geuenich2024impacts}. Moreover, the reliance on extensive annotations limits the diversity of available training data, particularly for rare diseases, where labeled samples are scarce~\cite{taherdoost2024beyond}. This constraint hampers the scalability of supervised learning methods and diminishes their ability to generalize across a wide spectrum of diseases~\cite{yoon2023domain}.

Recently, zero-shot learning (ZSL) based on vision-language alignment has emerged as a promising alternative to traditional supervised learning methods~\cite{long2017zero,zhang2024language}. By eliminating the need for extensive manual annotations, ZSL offers a more efficient path to diagnose a diverse range of abnormalities. For instance, CLIP~\cite{radford2021learning} leverages contrastive learning on large-scale image-text datasets to enable effective vision-language alignment and perform zero-shot classification without extensive manual annotations~\cite{zhao2023clip}. In the medical domain, several studies have demonstrated the potential of CLIP for zero-shot abnormality diagnosis on 3D CT volumes~\cite{hamamci2024foundation,cao2024bootstrapping,blankemeier2024merlin,shui2025large}. \textcolor{black}{However, directly applying CLIP to 3D medical images remains inadequate for vision–language alignment due to a significant modality gap between image and text embeddings. As shown in Fig.~\ref{fig:motivation}(a,b), image and text features form two distinct clusters even after alignment, with the feature alignment score (FAS) improving only from 0.08 to 0.37. This persistent separation indicates that standard CLIP fails to fully capture the semantic correspondence between visual and textual representations in the medical domain. Similar modality discrepancies have also been observed in previous studies~\cite{schrodi2024two,eslami2024mitigate}.} This highlights the necessity of addressing the modality gap in CLIP embeddings to achieve better alignment in 3D medical images.

In this paper, we propose a \textit{simple yet effective} framework named {\bf Bridged Semantic Alignment (\algname)},
which aims to bridge the modality gap and facilitate effective vision-language alignment. 
Our approach consists of two modules: semantic summarization and cross-modal knowledge interaction (CMKI). First, we utilize a large language model (LLM), which has powerful semantic understanding capabilities,  to summarize clinical reports and extract critical information. This significantly reduces the learning difficulty on the textual side and produces high-level semantic features. Next, we propose a CMKI module to supplement the conventional CLIP alignment, where a cross-modal knowledge bank (CMKB) acts as a semantic bridge between image and text features. By projecting image and text features into the shared latent space of the CMKB, it
reduces the discrepancy between modalities and simultaneously preserves their unique characteristics. This enables an implicit alignment approach that does not rely on paired image-text data. Finally, we employ contrastive learning to compare positive and negative pairs, achieving explicit vision-language alignment. Fig.~\ref{fig:motivation}(c) visualizes CMKB serving as a bridge between image and text features, effectively narrowing the modality gap. Also, the FAS is further improved to 0.50, indicating a better alignment.

To explore the potential of our vision-language alignment approach for zero-shot diagnosis, we expand the original 18 abnormality labels on CT-RATE~\cite{hamamci2024foundation} with 15 additional labels. The new benchmark, referred to as ``CT-RATE-LT", focuses on underrepresented abnormalities with limited occurrences in the dataset, providing a challenging yet valuable testing ground for evaluating diagnostic methods. Experimental results on CT-RATE-LT demonstrate that our method achieves excellent performance in diagnosing underrepresented abnormalities, showing significant improvements compared to the existing state-of-the-art (SOTA) method (AUC: 76.9 vs. 85.6). Furthermore, we evaluate our method on the benchmark  CT-RATE~\cite{hamamci2024foundation} and RAD-ChestCT~\cite{draelos2021machine} datasets, achieving SOTA performances in zero-shot abnormality diagnosis, with an AUC of 79.2 and 70.0 on internal and external validation, respectively. In addition to its diagnostic capabilities, our method significantly outperforms the SOTA in the report-to-volume retrieval task (Recall@10: 5.0 vs. 10.1), demonstrating effective semantic alignment between image and text features.

The main contributions of this work are as follows:
\begin{itemize}
\item We propose a \algname framework, consisting of semantic report summarization and cross-modal knowledge
interaction, which effectively acts as a bridge between the visual and texture features embeddings and hence forges a better vision-and-language alignment.
\item We introduce an expanded benchmark dataset, named ``CT-RATE-LT'', for 3D medical image analysis, encompassing 15 underrepresented abnormalities, which provides an effective tool for evaluating the zero-shot diagnosis capabilities on long-tailed minor diseases.
\item Our method achieves SOTA performance on zero-shot tasks across both internal and external validation datasets, demonstrating its effectiveness in diagnosing various abnormalities and its strong capability in the report-to-volume retrieval task.

\end{itemize}

\section{Related Work}
\label{sec:related_work}
 
\subsection{Medical Vision-language Pretraining}
\label{subsec:MVLP}
Existing medical vision-language pretraining (VLP) research is predominantly focused on 2D imaging. Public datasets like MIMIC-CXR~\cite{johnson2019mimic} have provided a solid foundation for the advancement of 2D VLP, while PadChest~\cite{bustos2020padchest}, which encompasses 193 abnormality categories, has established an effective benchmark for rare disease diagnosis. GLoRIA~\cite{huang2021gloria}  enhances the ability to capture associations between images and text by integrating global and local feature alignment. CheXZero~\cite{tiu2022expert} leverages a CLIP model pre-trained on natural data, achieving stable improvements in medical VLP performance. MedCLIP~\cite{wang2022medclip} employs unpaired medical images and text with a semantic matching loss to mitigate false negatives. Xplainer~\cite{pellegrini2023xplainer} introduces an explainable zero-shot diagnosis framework based on observation-driven contrastive learning. MedKLIP~\cite{wu2023medklip} and KAD~\cite{zhang2023knowledge} incorporate domain-specific medical knowledge to improve VLP performance in CXR diagnosis tasks. CARZero~\cite{lai2024carzero} advances cross-modal alignment by leveraging cross-attention mechanisms to address the complex relationships between visual and textual modalities. These advancements have established 2D VLP as a robust framework for various abnormalities diagnosis.

Recently, many researches have begun extending VLP to 3D medical images. CT-CLIP~\cite{hamamci2024foundation} combines spatial and causal transformers in 3D vision encoder. BIUD~\cite{cao2024bootstrapping} enhances 3D CT performance by distilling knowledge from CXR models. CT-GLIP~\cite{lin2024ct} and fVLM~\cite{shui2025large} propose anatomical structure alignment strategies to achieve organ-level feature alignment. Merlin~\cite{blankemeier2024merlin} introduces a two-step process, first optimizing the visual encoder with supervised labels and then training for vision-language alignment. Despite these advances, previous methods have overlooked the significant modality gap in CLIP embedding spacing for 3D medical imaging-report alignment, limiting the performance of zero-shot abnormality diagnosis.

\subsection{Multi-Modality Interaction}
\label{subsec:MI}

Cross-modal interaction has been extensively explored to bridge the gaps between modalities, such as vision and language modalities. Transformer-based approaches like ViLBERT~\cite{lu2019vilbert} employs dual-stream transformers with co-attentional layers for modality-specific processing, while ViLT~\cite{kim2021vilt} uses a unified transformer to jointly encode visual patches and textual embeddings, enabling efficient cross-modal interaction. MGCA~\cite{wang2022multi} introduces cross-attention mechanisms to achieve cross-modal interaction, aligning medical image and text representations at multiple levels. MPMA~\cite{zhang2023multi} enhances cross-modal interaction by integrating image-text reconstruction with a global and local alignment mechanism, enabling richer semantic representation learning. However, these approaches primarily emphasize explicit alignment mechanisms, often overlooking the potential of leveraging shared semantic spaces to guide cross-modal learning.

Dictionary learning has emerged as a technique for enhancing cross-modal alignment by constructing shared representational spaces.
Deng et al.~\cite{deng2015discriminative} aligns features via a shared label space to improve cross-modal retrieval.
HCDDL~\cite{li2023hierarchical} builds hierarchical semantic embeddings by dictionary learning for fine-grained alignment.
UNIMO-2~\cite{li2022unimo} leverages grounded semantic spaces to bridge visual and textual modalities, effectively addressing misalignment.
Methods such as DetCLIP~\cite{yao2022detclip} and SOHO~\cite{huang2021seeing} further exploit trainable dictionaries to enrich visual-semantic representations, achieving SOTA results in zero-shot and vision-language tasks.
CP-CLIP~\cite{yu2024core} employs the Core–Periphery principle to facilitate cross-modal interaction by structuring shared semantics and modality-specific features within a unified latent space.
\textcolor{black}{
Despite these advances, prior dictionary-based methods mainly rely on reconstruction losses, lacking discriminative constraints and risking cross-modal leakage.
To overcome these issues, we design a Cross-Modal Knowledge Interaction (CMKI) module that jointly optimizes reconstruction and contrastive objectives within separate modality-specific banks.
This implicit–explicit synergy preserves modality integrity and ensures robust alignment for 3D medical imaging.}

\begin{figure*}[t]
  \centering
    \includegraphics[width=0.95\linewidth]{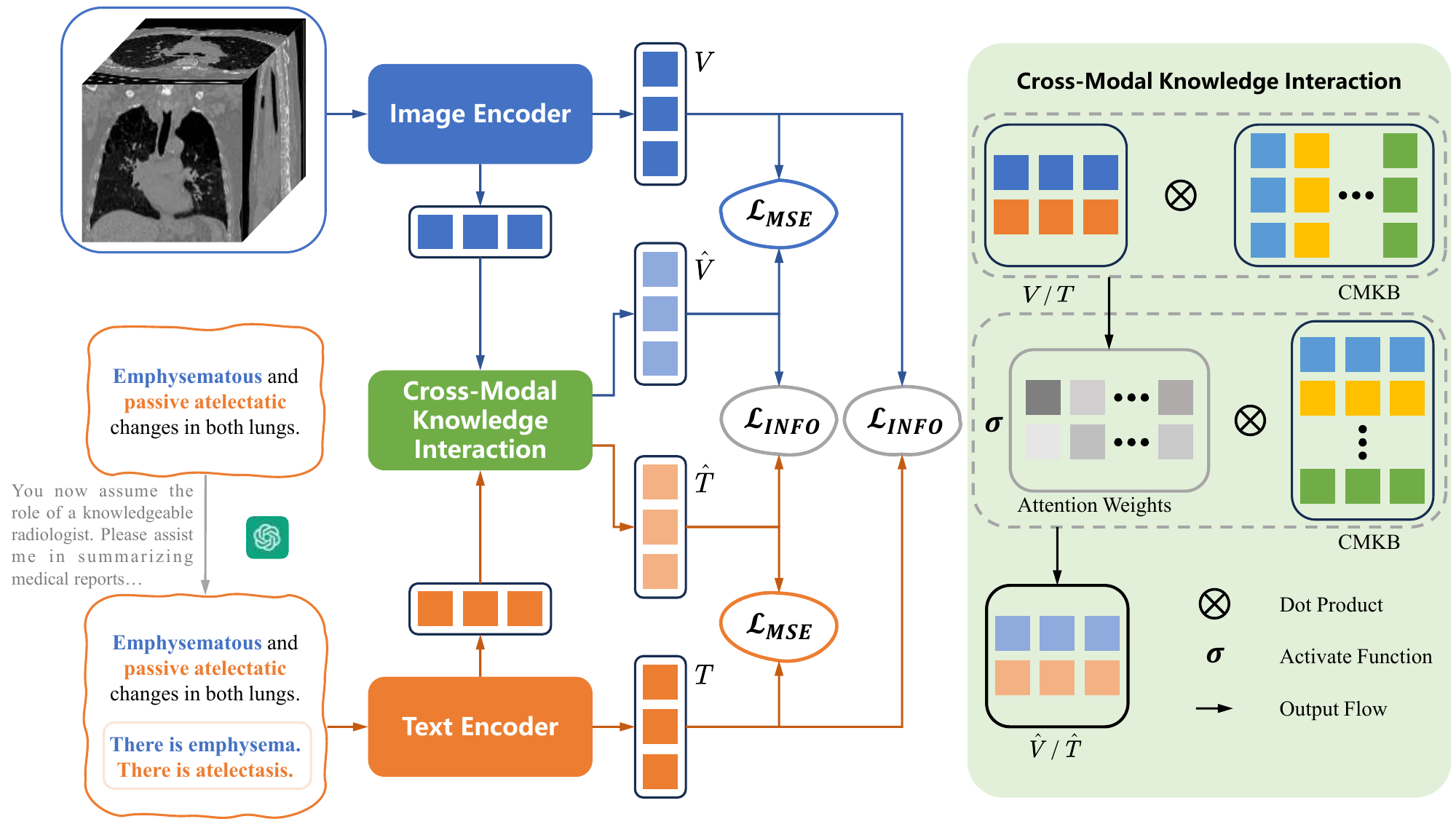}
    \caption{\textcolor{black}{Illustration of the proposed \algname network, which integrates semantic summarization and cross-modal knowledge interaction (CMKI). First, we leverage a large language model (LLM) to summarize the report, generating outputs in a fixed template. These summarized reports, along with the original reports, serve as the textual inputs. Then, image and text features are extracted by respective encoders and fed into CMKI module to obtain interaction features. Finally, the interaction features are constrained using an MSE loss, while alignment optimization is achieved via an InfoNCE loss.}} 
    \label{fig:Network}
    \vspace{-4mm}
\end{figure*}

\section{Method}
\label{sec:method}

In this section, we present the proposed \algname framework for zero-shot classification, consisting of semantic summarization and a cross-modal knowledge interaction (CMKI) module (Fig.~\ref{fig:Network}).  Further details are provided below.

\subsection{Semantic Summarization}
\label{subsec:ss}
\begin{figure}[t]
  \centering
    \includegraphics[width=0.95\linewidth]{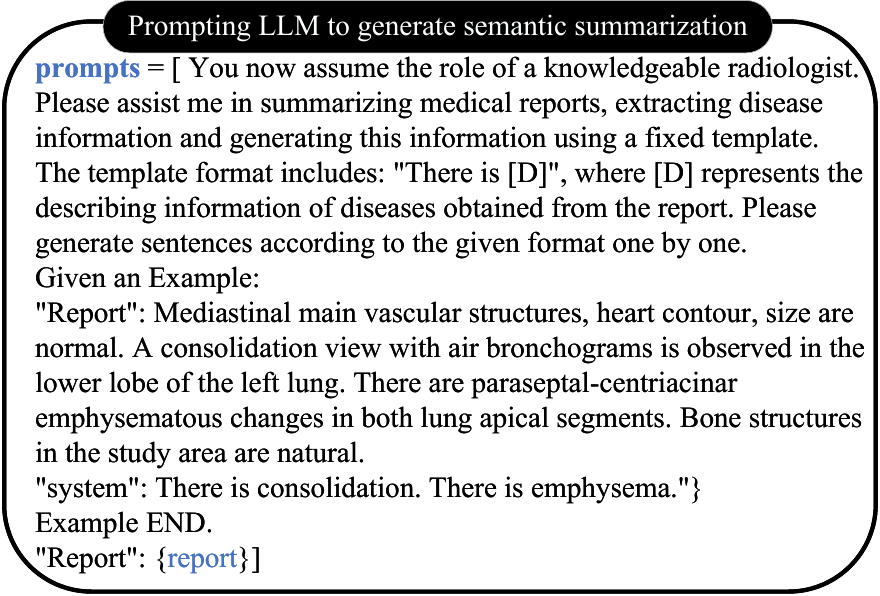}
    \caption{Prompt for LLM used in semantic summarization for reports.} 
    \label{fig:ppt}
    \vspace{-4mm}
\end{figure}

To simplify learning on the textual side, we introduce semantic summarization. In the context of medical reports, textual data often contains lengthy and complex descriptions, including both critical abnormality-related information and a significant amount of irrelevant details. Such data characteristics increase the difficulty of extracting and aligning core semantic information with image features. We leverage the powerful semantic understanding and summarization capabilities of LLM to produce semantic summaries of medical reports, which are much easier to understand. As illustrated in Fig.~\ref{fig:ppt}, a prompt is designed to guide the LLM in extracting key abnormality-related information and generating outputs in a fixed template: ``There is [abnormality].'' This template-based summarization significantly reduces the alignment difficulty by providing more focused and consistent descriptions.

However, solely relying on summarized reports may lead to information loss, especially when subtle or nuanced abnormalities are omitted. To balance simplicity and completeness, we propose a dual-input strategy that incorporates both the original report and its corresponding summary. This combined representation retains the rich contextual information from the full report while benefiting from the semantic clarity of the summary. As a result, it enhances the robustness and accuracy of image-text alignment in \algname training.

\subsection{Feature Extraction}
\label{subsec:FE}

Assume that the training dataset contains \( N \) samples denoted as \( S_{\text{train}} = \left\{\left(I_{1}, R_{1}\right), \ldots, \left(I_{N}, R_{N}\right)\right\} \), where \( I_{i} \in \mathbb{R}^{H \times W \times D} \) represents a 3D CT volume and \( R_{i} \) represents its corresponding medical report. \textcolor{black}{Specifically, \( R_{i} \) refers to the concatenation of the original medical report and its semantic summarization, which is the dual-input strategy. The combined text is then used for alignment with the image features.} Here, \( H, W, D \) denote the height, width, and depth of the 3D CT volumes, respectively. As illustrated in Fig.~\ref{fig:Network}, we introduce the components of our architecture for feature extraction, including an image encoder \( \mathbf{\Phi}_{\text{I}} \) for CT images and a text encoder \( \mathbf{\Phi}_{\text{R}} \) for medical reports.

\subsubsection{Image Encoder} The image encoder is designed to extract global features from 3D CT images, as shown in Eq.~\ref{img_encoder}.
\begin{equation}
\label{img_encoder}
V_i = \mathbf{\Phi}_{\text{I}}\left(I_{i}\right),
\end{equation}
where \( V_i \in \mathbb {R}^{d} \) represents the global features of the image, and \( d \) denotes the feature dimension. In our experiments, we adopt a ViT-base~\cite{dosovitskiy2020image} model as the image encoder. \textcolor{black}{The output of the \texttt{[CLS]} token is used to represent the global feature for 3D image, following the methodology in~\cite{dosovitskiy2020image}.}

\subsubsection{Text Encoder} The text encoder is used to extract global features from the medical report, as shown in Eq.~\ref{txt_encoder}.
\begin{equation}
\label{txt_encoder}
T_i = \mathbf{\Phi}_{\text{R}}\left(R_{i}\right),
\end{equation}
where \( T_i \in \mathbb {R}^{d} \) represents the global features of the medical report. \textcolor{black}{We use CXRBERT~\cite{boecking2022making} as the text encoder, which is specifically pretrained on large-scale radiology reports and thus better suited for chest imaging tasks.} \textcolor{black}{The output of the \texttt{[CLS]} token is used as the global feature for the text~\cite{devlin2019bert}.} In our experiment, the output feature dimensions of both the image and text encoders are unified to \( d\).

\subsection{Cross-Modal Knowledge Interaction}
\label{subsec:CMKI}

To bridge the gap between 3D medical images and textual reports and enhance vision-language alignment, we propose the CMKI module, as shown in Fig.~\ref{fig:Network}. This module effectively reduces cross-modal discrepancies, improves inter-modal similarity, and preserves the unique characteristics of each modality, thereby enhancing the overall quality of vision-language alignment.

In the CMKI module, we introduce a cross-modal knowledge bank (CMKB) as a bridge for image-text interaction. The CMKB is initialized as a set of learnable embeddings:
\begin{equation}
\mathbf{B} = \{ \mathbf{b}_1, \mathbf{b}_2, \dots, \mathbf{b}_K \}, \quad \mathbf{b}_k \in \mathbb{R}^{d},
\end{equation}
where each \(\mathbf{b}_k\) denotes a basis vector and \( K \) represents the number of basis vectors.

To construct a bridge between the image and text features, we first calculate the similarity between the image features \( V_i \) and the CMKB basis vectors \( \mathbf{B} \). This similarity is calculated using the inner product:
\begin{equation}
\mathrm{sim}(\mathbf{B}, V_i) = \mathbf{B}^\top V_i,
\end{equation}
where \(\mathrm{sim}(\mathbf{B}, V_i)\) represents the similarity between the image feature \( V_i \) and each basis vector in the CMKB. The similarity scores are then normalized using a softmax function to produce attention weights:
\begin{equation}
\mathbf{Z}_i^{V} = \mathrm{softmax}\left(\mathrm{sim}(\mathbf{B}, V_i)\right).
\end{equation}
Here, \( \mathbf{Z}_i^{V} \) denotes the attention weights that highlight the contribution of each basis vector to the reconstruction of \( V_i \). Similarly, for the text features \( T_i \), the attention weights \( \mathbf{Z}_i^{T} \) is calculated as:
\begin{equation}
\mathbf{Z}_i^{T} = \mathrm{softmax}\left(\mathrm{sim}(\mathbf{B},T_i)\right).
\end{equation}
These attention weights are used to identify the most relevant basis vectors from the CMKB for representing the image and text features.

\textcolor{black}{Using the attention weights \( \mathbf{Z}_i^{V} \) and \( \mathbf{Z}_i^{T} \), the original image and text features are represented in the CMKB space by taking a weighted combination of the basis vectors, yielding their approximated representations:}
\begin{equation}
\hat{V}_i = \mathbf{B} \cdot \mathbf{Z}_i^{V},~~\hat{T}_i = \mathbf{B} \cdot \mathbf{Z}_i^{T}.
\end{equation}
In this process, the CMKB captures the common features shared by the image and text modalities, enabling cross-modal information exchange.

We use a reconstruction loss to optimize the CMKB. The loss function is defined as:
\begin{equation}
\ell_{\text{MSE}} = \frac{1}{N} \sum_{i=1}^{N} \left(\| V_i - \hat{V}_i \|_2^2 + \| T_i - \hat{T}_i \|_2^2 \right).
\end{equation}
The reconstruction mechanism ensures that the CMKB functions as an effective bridge for image-text interaction by capturing critical information from both modalities, facilitating cross-modal knowledge exchange, and reducing modality discrepancies. By narrowing the modality gap, the CMKI module enhances the alignment between 3D CT volumes and textual reports, thereby improving the effectiveness of CLIP's training. Specifically, our method minimizes cross-modal discrepancies, allowing CLIP to more effectively reduce the distance between paired image-text features and increase the distance between unpaired ones during contrastive learning. Furthermore, the use of CMKB as an intermediary maintains the independence of image and text features, preventing information leakage between positive and negative pairs in contrastive learning. This design ensures robust and unbiased feature learning while enhancing vision-language alignment.

To ensure vision-language alignment, we employ a symmetric InfoNCE loss between image and text features:
\begin{equation}
\ell_{\text{INFO}} = -\frac{1}{N} \sum_{i=1}^{N} \log \frac{\exp\left(\text{sim}(V_i, T_i) / \tau\right)}{\sum_{j} \exp\left(\text{sim}(V_i, T_j) / \tau\right)},
\end{equation}
where \( \text{sim}(\cdot, \cdot) \) denotes cosine similarity, and \( \tau \) is a temperature parameter. Additionally, we apply the same InfoNCE loss to the reconstructed features:
\begin{equation}
\ell_{\text{INFO-R}} = -\frac{1}{N} \sum_{i=1}^{N} \log \frac{\exp\left(\text{sim}(\hat{V}_i, \hat{T}_i) / \tau\right)}{\sum_{j} \exp\left(\text{sim}(\hat{V}_i, \hat{T}_j) / \tau\right)}.
\end{equation}

The total alignment loss combines the reconstruction and InfoNCE losses:
\begin{equation}
\ell_{\text{total}} = \alpha \ell_{\text{MSE}} + \beta \ell_{\text{INFO}} + \gamma \ell_{\text{INFO-R}},
\end{equation}
where \(\alpha\), \(\beta\), and \(\gamma\) are hyperparameters controlling the contributions of the reconstruction loss (\(\ell_{\text{MSE}}\)), the InfoNCE loss (\(\ell_{\text{INFO}}\)), and the InfoNCE loss on reconstructed features (\(\ell_{\text{INFO-R}}\)), respectively. \textcolor{black}{To further clarify the computational flow, we provide a concise pseudocode implementation of the CMKI module in Algorithm~\ref{alg:cmki}.}

\begin{algorithm}[t]
\caption{\textcolor{black}{Cross-Modal Knowledge Interaction}}
\label{alg:cmki}
\begin{algorithmic}[1]
\Require Image features $V_i$, text features $T_i$, CMKB $B$, temperature $\tau$
\State $S_V \gets B^\top V_i$, \quad $S_T \gets B^\top T_i$
\State $Z_i^V \gets \mathrm{softmax}(S_V)$, \quad $Z_i^T \gets \mathrm{softmax}(S_T)$
\State $\hat V_i \gets B Z_i^V$, \quad $\hat T_i \gets B Z_i^T$
\State $\ell_{\text{MSE}} \gets \|V_i-\hat V_i\|^2 + \|T_i-\hat T_i\|^2$
\State $\ell_{\text{INFO}} \gets -\log \frac{\exp(\mathrm{sim}(V_i,T_i)/\tau)}{\sum_j \exp(\mathrm{sim}(V_i,T_j)/\tau)}$
\State $\ell_{\text{INFO-R}} \gets -\log \frac{\exp(\mathrm{sim}(\hat V_i,\hat T_i)/\tau)}{\sum_j \exp(\mathrm{sim}(\hat V_i,\hat T_j)/\tau)}$
\State $\ell_{\text{total}} \gets \alpha\ell_{\text{MSE}} + \beta\ell_{\text{INFO}} + \gamma\ell_{\text{INFO-R}}$
\end{algorithmic}
\end{algorithm}

\begin{figure}[t]
 \centering{
    \includegraphics[width=0.95\linewidth]{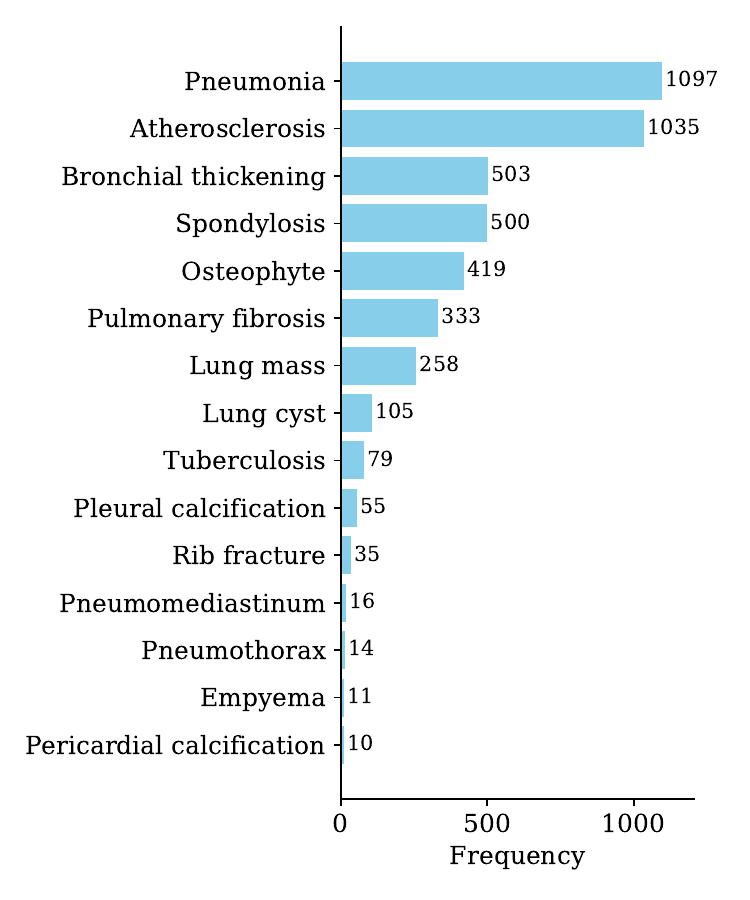}
  }
  \caption{Histogram of abnormality frequencies for CT-RATE-LT.} 
  \label{fig:dfh}
  \vspace{-4mm}
\end{figure}

\begin{figure}[th]
 \centering{
    \includegraphics[width=0.95\linewidth]{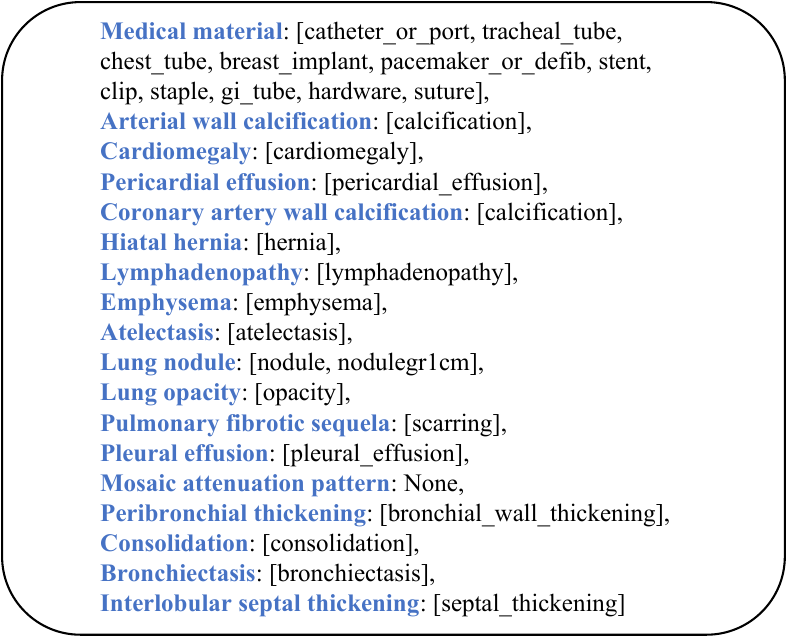}
  }
  \caption{Mapping of 27 abnormalities from RAD-ChestCT to 18 abnormalities in CT-RATE. The abnormalities from CT-RATE are denoted in {\color{blue}blue font}, whereas the abnormalities from RAD-ChestCT are denoted in black font.
} 
  \label{fig:dmp}
  \vspace{-4mm}
\end{figure}

\section{Experiments}
\label{sec:experiments}

\subsection{Materials}

We conduct experiments on three publicly available datasets: CT-RATE~\cite{hamamci2024foundation}, RAD-ChestCT~\cite{draelos2021machine}, and INSPECT~\cite{huang2023inspect}. Detailed descriptions of these datasets are provided below. 

\textbf{CT-RATE}~\cite{hamamci2024foundation}: The CT-RATE dataset comprises 25,692 non-contrast chest CT volumes, expanded to 50,188 volumes through various reconstruction techniques, originating from 21,304 unique patients. Each CT volume is paired with a corresponding radiology report. A total of 47,149 CT volumes with corresponding reports are used for training, while 3,039 CT volumes are reserved for internal validation. The dataset includes 18 abnormality labels, extracted from reports using an automated algorithm. 

\textcolor{black}{To further evaluate the zero-shot detection capabilities of our method, we construct an extended \textbf{CT-RATE-LT} dataset focusing on long-tail abnormalities. Specifically, we first predefine a set of 15 underrepresented abnormalities based on} \textcolor{black}{their clinical relevance and low prevalence in CT-RATE. GPT-4 Turbo~\footnote{\url{https://openai.com/index/gpt-4/}} is then prompted to determine the presence or absence of these abnormalities in the impression sections of the reports. All GPT-4 annotations are subsequently verified by a board-certified radiologist, who carefully compares each predicted label with the original report description. Inconsistent or ambiguous cases are corrected based on clinical context, ensuring that each abnormality label precisely reflects the radiology report.} The frequency distribution of CT-RATE-LT abnormalities is shown in Fig.~\ref{fig:dfh}.

\textbf{RAD-ChestCT}~\cite{draelos2021machine}: The RAD-ChestCT dataset contains 3,630 CT volumes, uniformly reconstructed using a single technique. This dataset includes 84 abnormality labels. As part of our external validation, as shown in Fig.~\ref{fig:dmp}, 27 abnormalities from RAD-ChestCT are mapped to the 18 abnormalities in CT-RATE to ensure consistency. To further demonstrate the strength of our method, we refer to the remaining 56 abnormalities from RAD-ChestCT as \textbf{RAD-ChestCT-LT}~\cite{draelos2021machine} and evaluate the zero-shot diagnosis performance on these additional abnormalities. Note that the abnormality, ``other\_path'',  is excluded from our analysis due to its lack of meaningful semantic descriptions.

\textcolor{black}{\textbf{INSPECT}~\cite{huang2023inspect}: We use the official test set comprising 3,214 contrast-enhanced CT pulmonary angiography (CTPA) cases to evaluate model generalization on the pulmonary embolism (PE) diagnosis task.}

\subsection{Dataset Preprocessing}
\label{sec:DP}

For the image data, we unify CT volume spacing to \(1.5\times1.5\times3\,\mathrm{mm}\) for uniform resolution across datasets. Random cropping during training generates input volumes of shape \(224\times224\times112\), balancing computational efficiency and spatial coverage for 3D images. The ViT-Base patch size is \(16\times16\times8\) for 3D inputs. To enhance feature representation, CT images are clipped to \([-1000,1000]\) HU to focus on clinically relevant tissues and normalized to \([-1,1]\).

\textcolor{black}{For the text data, we process both the original reports and the semantic summaries generated by GPT-4 Turbo via the OpenAI API without additional fine-tuning, splitting them into shorter sentences for analysis.} As part of our data augmentation strategy, five sentences are randomly sampled and concatenated into a longer sentence during training, enabling diverse text representations that better capture the variability in medical reporting styles.

\begin{table*}
    \centering
    \setlength{\tabcolsep}{3pt}
    \renewcommand{\arraystretch}{1.}
    \scriptsize
    \caption{\textcolor{black}{Zero-shot abnormality diagnosis performance comparison across internal (CT-RATE) and external (RAD-ChestCT) validation benchmarks. $^{*}$ indicates that statistical significance ($p < 0.001$) is confirmed for \algname against other methods.}}
    \begin{tabular}{@{}l *{7}{c} *{7}{c}@{}}
    \toprule
    \multirow{2}{*}{Method} & \multicolumn{7}{c}{CT-RATE} & \multicolumn{7}{c}{RAD-ChestCT} \\
    \cmidrule(lr){2-8} \cmidrule(lr){9-15}
      & AUC & ACC & F1  & Pre & mAP & Rec@1 & Pre@3
      & AUC & ACC & F1  & Pre & mAP & Rec@1 & Pre@3 \\
    \midrule
    CT-Net~\cite{draelos2021machine}$^{*}$ 
      & 60.3$\pm$1.2 & 58.1$\pm$0.9 & 63.1$\pm$0.8 & 23.9$\pm$1.2 & 30.0$\pm$1.6 & 11.5$\pm$0.4 & 24.6$\pm$0.5
      & 54.4$\pm$1.2 & 54.0$\pm$0.8 & 58.7$\pm$0.7 & 28.5$\pm$1.0 & 32.1$\pm$1.3 & 13.6$\pm$0.3 & 36.4$\pm$0.4 \\
    VocabFine~\cite{hamamci2024foundation}$^{*}$ 
      & 75.6$\pm$1.2 & 70.5$\pm$0.8 & 73.8$\pm$0.7 & 35.3$\pm$1.4 & 42.8$\pm$2.0 & 17.3$\pm$0.5 & 35.3$\pm$0.6
      & 65.0$\pm$1.2 & 61.5$\pm$0.8 & 65.9$\pm$0.7 & 34.9$\pm$1.1 & 38.9$\pm$1.4 & 14.6$\pm$0.3 & 41.8$\pm$0.5 \\
    ClassFine~\cite{hamamci2024foundation}$^{*}$ 
      & 75.6$\pm$1.1 & 68.9$\pm$0.8 & 72.4$\pm$0.7 & 33.9$\pm$1.4 & 39.9$\pm$2.0 & 16.7$\pm$0.5 & 32.4$\pm$0.5
      & 64.3$\pm$1.3 & 60.7$\pm$0.8 & 64.9$\pm$0.7 & 34.9$\pm$1.1 & 37.6$\pm$1.4 & 14.6$\pm$0.3 & 36.7$\pm$0.5 \\
    CT-CLIP~\cite{hamamci2024foundation}$^{*}$ 
      & 73.1$\pm$1.1 & 66.8$\pm$0.8 & 70.7$\pm$0.7 & 32.3$\pm$1.3 & 36.1$\pm$1.7 & 7.9$\pm$0.3 & 26.2$\pm$1.0
      & 62.9$\pm$1.3 & 59.5$\pm$0.8 & 64.2$\pm$0.7 & 33.6$\pm$1.0 & 33.9$\pm$1.3 & 9.6$\pm$0.2 & 32.9$\pm$0.4 \\
    BIUD~\cite{cao2024bootstrapping} 
      & 71.3 & 68.1 & 71.6 & 33.8 & -- & -- & --
      & 62.9 & 60.6 & 65.2 & 33.7 & -- & -- & -- \\
    Merlin~\cite{blankemeier2024merlin}$^{*}$ 
      & 72.8$\pm$1.3 & 67.2$\pm$0.9 & 70.9$\pm$0.8 & 33.7$\pm$1.2 & 34.1$\pm$1.6 & 8.1$\pm$0.3 & 25.0$\pm$0.5
      & 64.4$\pm$1.2 & 61.9$\pm$0.8 & 66.3$\pm$0.7 & 34.8$\pm$1.0 & 35.7$\pm$1.2 & 10.7$\pm$0.3 & 35.9$\pm$0.4 \\
    DCFormer~\cite{ates2025dcformer}$^{*}$ 
      & 67.0$\pm$1.2 & 62.4$\pm$0.9 & 66.8$\pm$0.8 & 28.6$\pm$1.2 & 30.2$\pm$1.6 & 12.2$\pm$0.4 & 24.9$\pm$0.5
      & 58.1$\pm$1.3 & 55.8$\pm$0.8 & 60.8$\pm$0.8 & 29.9$\pm$1.0 & 31.4$\pm$1.2 & 14.4$\pm$0.3 & 33.8$\pm$0.4 \\
    X2CT-CLIP~\cite{you2025x2ct} 
      & 71.6 & -- & -- & -- & -- & -- & --
      & 64.5 & -- & -- & -- & -- & -- & -- \\
    fVLM~\cite{shui2025large}$^{*}$ 
      & 77.8$\pm$1.1 & 71.8$\pm$0.8 & 75.1$\pm$0.7 & 37.9$\pm$1.4 & 47.6$\pm$2.1 & 16.0$\pm$0.5 & 33.8$\pm$0.6
      & 68.0$\pm$1.2 & 64.7$\pm$0.8 & 68.8$\pm$0.7 & 37.4$\pm$1.1 & 43.5$\pm$1.5 & 13.7$\pm$0.3 & 41.1$\pm$0.5 \\
    \algname 
      & \textbf{82.9$\pm$1.0} & \textbf{77.0$\pm$0.7} & \textbf{79.3$\pm$0.7} & \textbf{43.2$\pm$1.5} & \textbf{55.4$\pm$2.3} & \textbf{21.4$\pm$0.5} & \textbf{43.2$\pm$0.6}
      & \textbf{74.2$\pm$1.0} & \textbf{68.6$\pm$0.8} & \textbf{72.0$\pm$0.7} & \textbf{42.2$\pm$1.2} & \textbf{49.4$\pm$1.6} & \textbf{14.9$\pm$0.3} & \textbf{48.7$\pm$0.5} \\
    \bottomrule
    \end{tabular}
    \label{tab:comparison}
    \vspace{-4mm}
\end{table*}

\subsection{Implementation Details}
\label{sec:ID}

In our experiments, we adapt ViT-B/16~\cite{dosovitskiy2020image} as the image encoder by replacing the 2D convolutions in the patch-embedding module with 3D convolutions, yielding a 3D ViT-B/16. For the text encoder, we fine-tune CXRBERT~\cite{boecking2022making} on CT-RATE reports to capture medical-domain language. We also use M3AE~\cite{chen2022m3ae} to pretrain both encoders. M3AE employs masked cross-modal reconstruction, where masked regions in one modality are reconstructed from the other, enhancing image–text semantic consistency and alignment. \textcolor{black}{The CMKB basis vectors are initialized using orthogonal initialization.}

In our experiments, $\alpha$, $\beta$, and $\gamma$ are empirically set to 0.5, 1, and 1, respectively, based on preliminary trials to balance the contributions of different loss components. $K$ is set to 2048, as determined through the experimental search results presented later. We use the Adam optimizer with a learning rate of $5 \times 10^{-5}$. \textcolor{black}{For the ablation studies, the batch size is set to 8 due to computational cost considerations, while our final configuration adopts a batch size of 64, which yields the best overall performance.} The code is implemented using the PyTorch framework, and all experiments are conducted on a single NVIDIA A800 GPU. \textcolor{black}{During the inference phase, we perform zero-shot abnormality diagnosis by setting the prompt template as ``There is [abnormality]'' and computing the similarity score for each category as
\[
s_c = \tfrac{1}{2}\big(\text{sim}(V, T_c) + \text{sim}(\hat{V}, T_c)\big).
\]
where $s_c$ denotes the final similarity score between the image and category $c$. For cross-modal retrieval tasks, only the original reports are used as queries, and no additional summarization step is required, ensuring efficient inference.}

\subsection{Evaluation Metrics}

\textcolor{black}{For the multi-label classification, we evaluate zero-shot abnormality diagnosis using Area under the ROC Curve (AUC), Accuracy (ACC), F1 score (F1), and Precision (Pre). 
To better capture clinical usefulness, we additionally report mean Average Precision (mAP), Recall@1 (Rec@1), and Precision@3 (Pre@3), where Recall@1 and Precision@3 reflect report-level retrieval performance.} 
Following fVLM~\cite{shui2025large}, we determine the threshold for F1 and Precision by searching the value that minimizes the Euclidean distance to the ideal point (FPR=0, TPR=1) on the ROC curve, representing the optimal trade-off between sensitivity and specificity. 
For the volume-to-volume retrieval task, we use Mean Average Precision at $Q$ (MAP@$Q$) metrics, where $Q$ represents the number of top-ranked retrieved volumes considered. In our experiment, $Q$ is set to $\{5, 10, 50\}$. 
For the report-to-volume retrieval task, we use Recall@$P$, where $P$ represents the number of retrieved images, to measure the proportion of correctly matched images within the top-$P$ retrieved results for a given query text. In our experiment, $P$ is set to $\{5, 10, 50, 100\}$.
\textcolor{black}{To assess statistical significance, we apply per-class DeLong tests for AUC, paired bootstrap resampling for ACC, F1, Precision, and mAP, and paired $t$-tests for Rec@1 and Pre@3. 
For retrieval metrics (MAP@$Q$ and Recall@$P$), statistical significance is assessed over queries using paired bootstrap. Throughout the tables, $^{*}$ indicates that statistical significance ($p < 0.001$) is confirmed for \algname against other methods.}

\begin{table*}[h]
    \setlength{\tabcolsep}{2.8pt}
    \centering
    \renewcommand{\arraystretch}{1.}
    \scriptsize
    \caption{\textcolor{black}{Performance comparison for zero-shot abnormality diagnosis on CT-RATE-LT and RAD-ChestCT-LT datasets.}}
    \begin{tabular}{@{}l *{7}{c} *{7}{c}@{}}
    \toprule
    \multirow{2}{*}{Method} & \multicolumn{7}{c}{CT-RATE-LT} & \multicolumn{7}{c}{RAD-ChestCT-LT} \\
    \cmidrule(lr){2-8} \cmidrule(lr){9-15}
      & AUC & ACC & F1  & Pre & mAP & Rec@1 & Pre@3
      & AUC & ACC & F1  & Pre & mAP & Rec@1 & Pre@3 \\
    \midrule

    VocabFine~\cite{hamamci2024foundation}$^{*}$
      & 71.5$\pm$2.8 & 67.3$\pm$0.8 & 75.4$\pm$0.7 & 16.3$\pm$0.8 & 19.6$\pm$1.9 & 15.1$\pm$0.6 & 15.1$\pm$0.4
      & 59.0$\pm$2.2 & 57.5$\pm$0.8 & 66.6$\pm$0.7 & 13.4$\pm$0.7 & 14.7$\pm$1.1 & 2.2$\pm$0.1 & 13.8$\pm$0.3 \\
    CT-CLIP~\cite{hamamci2024foundation}$^{*}$
      & 67.9$\pm$3.2 & 65.7$\pm$0.9 & 74.0$\pm$0.7 & 15.4$\pm$0.8 & 17.1$\pm$1.4 & 6.2$\pm$0.4 & 10.2$\pm$0.3
      & 58.2$\pm$2.2 & 56.9$\pm$0.8 & 66.2$\pm$0.7 & 13.1$\pm$0.7 & 13.9$\pm$1.0 & 1.9$\pm$0.1 & 10.0$\pm$0.3 \\
    Merlin~\cite{blankemeier2024merlin}$^*$ 
      & 67.0$\pm$3.1 & 62.5$\pm$0.9 & 71.4$\pm$0.7 & 14.5$\pm$0.8 & 16.1$\pm$1.4 & 7.2$\pm$0.4 & 11.7$\pm$0.3
      & 59.1$\pm$2.2 & 57.0$\pm$0.8 & 66.1$\pm$0.7 & 13.6$\pm$0.7 & 15.3$\pm$1.2 & 3.2$\pm$0.1 & 18.3$\pm$0.4 \\
    \algname
      & \textbf{81.5$\pm$2.0} & \textbf{75.8$\pm$0.8} & \textbf{81.8$\pm$0.6} & \textbf{20.2$\pm$1.0} & \textbf{26.8$\pm$0.2} & \textbf{19.6$\pm$0.7} & \textbf{19.4$\pm$0.4}
      & \textbf{69.0$\pm$2.0} & \textbf{65.9$\pm$0.8} & \textbf{73.3$\pm$0.6} & \textbf{17.7$\pm$0.9} & \textbf{22.2$\pm$1.7} & \textbf{4.0$\pm$0.1} & \textbf{23.6$\pm$0.4} \\
    \bottomrule
    \end{tabular}
    \label{tab:comparison_combined}
    \vspace{-4mm}
\end{table*}

\begin{table}[t]
    \setlength{\tabcolsep}{6pt}
    \centering
    \renewcommand{\arraystretch}{1.0}
    \scriptsize
    \caption{\textcolor{black}{Performance comparison for zero-shot diagnosis on the INSPECT dataset.}}
    \begin{tabular}{@{}lccccc@{}}
    \toprule
    Method & AUC & ACC & F1 & Pre & mAP \\
    \midrule
    VocabFine~\cite{hamamci2024foundation}$^{*}$ 
      & 50.2$\pm$1.3 & 50.6$\pm$0.9 & 55.1$\pm$0.9 & 21.6$\pm$1.0 & 21.7$\pm$1.1 \\
    CT-CLIP~\cite{hamamci2024foundation}$^{*}$   
      & 53.9$\pm$1.2 & 52.0$\pm$0.9 & 56.4$\pm$0.8 & 23.0$\pm$1.1 & 23.6$\pm$1.2 \\
    Merlin~\cite{blankemeier2024merlin}$^*$ 
      & 50.7$\pm$1.2 & 46.9$\pm$0.9 & 51.5$\pm$0.9 & 21.3$\pm$1.0 & 21.1$\pm$1.0 \\
    \algname 
      & \textbf{60.5$\pm$1.2} & \textbf{60.3$\pm$0.9} & \textbf{63.9$\pm$0.8} & \textbf{27.4$\pm$1.2} & \textbf{27.2$\pm$1.4} \\
    \bottomrule
    \end{tabular}
    \label{tab:inspect_binary}
    \vspace{-4mm}
\end{table}

\section{Results and Analysis}
\label{sec:ReAn}

\subsection{Comparison with State-of-the-art Methods}
\label{sec:CSOTA}

We compare our method against six zero-shot approaches, supervised learning, and fine-tuning methods, as summarized in Table~\ref{tab:comparison}. The evaluation is conducted on both internal (CT-RATE~\cite{hamamci2024foundation}) and external (RAD-ChestCT~\cite{draelos2021machine}) benchmarks for zero-shot abnormality diagnosis. To provide an evaluation, we detail the comparison methods as follows. Some methods do not report certain metrics in their original papers; we denote these missing values with ``--'' in the table.

\noindent
\textbf{Supervised Learning.} 
CT-Net~\cite{draelos2021machine} uses a deep convolutional neural network with transfer learning and 3D convolutions to perform multi-disease classification.  

\noindent
\textbf{Fine-tuning Methods.} 
VocabFine~\cite{hamamci2024foundation} fine-tunes only the projection layers of the image and text encoders in CT-CLIP while freezing all other pre-trained layers, preserving open-vocabulary capabilities. 
ClassFine~\cite{hamamci2024foundation} adds and trains a new classification layer on CT-CLIP while optionally freezing the pre-trained layers to retain existing feature representations.  

\noindent
\textbf{Zero-shot Methods.} 
CT-CLIP~\cite{hamamci2024foundation} employs a contrastive vision-language pretraining approach to align 3D chest CT volumes with radiology reports, utilizing a vision transformer for 3D CT and a text transformer for embedding generation. 
BIUD~\cite{cao2024bootstrapping} aligns 3D CT images with 2D X-ray images through a language-guided retrieval strategy and enhances alignment using contrastive learning and entity-focused masking. 
Merlin~\cite{blankemeier2024merlin} employs multi-task learning to process 3D abdominal CT volumes by integrating structured electronic health record data and unstructured radiology reports, optimizing binary cross-entropy and InfoNCE losses to achieve effective vision-language alignment. 
DCFormer~\cite{ates2025dcformer} proposes a lightweight 3D vision-language encoder using decomposed convolutions for efficient volumetric representation learning; integrated into a CLIP framework, it reduces computational cost while maintaining strong alignment between CT images and radiology reports. 
X2CT-CLIP~\cite{you2025x2ct} introduces a tri-modal contrastive learning framework that aligns CXRs with 3D CT volumes and CT reports in a shared latent space, transferring knowledge from pretrained CT encoders to a CXR encoder using simulated image triplets to enable CT-level multi-abnormality detection from CXRs. 
fVLM~\cite{shui2025large} utilizes fine-grained contrastive learning for anatomy-level alignment between CT images and radiology reports, addressing false negatives with a dual reduction strategy and employing co-teaching to improve semantic representation.

As shown in Table~\ref{tab:comparison}, our method achieves the highest evaluation metrics among all tested methods on both the CT-RATE and RAD-ChestCT datasets. Compared to supervised learning and fine-tuning methods, our approach demonstrates clear advantages, attributed to the proposed \algname framework for vision-language semantic alignment. While fVLM introduces fine-grained anatomy-level alignment to improve vision-language tasks, it overlooks the modality gap inherent in the CLIP embedding space, which limits its ability to fully align image and text features. In contrast, our \algname framework effectively narrows this gap by combining the semantic summarization module, which simplifies textual learning and extracts high-level semantic information, with the CMKI module, which optimizes image-text features by reducing their discrepancies and preserving their unique characteristics. This comprehensive approach ensures superior vision-language alignment and enables our method to achieve SOTA performance in zero-shot abnormality diagnosis. Surprisingly, our method achieves outstanding performance using low-resolution 3D images, further highlighting its robustness and practicality for resource-constrained clinical environments.

\textcolor{black}{To comprehensively evaluate the zero-shot diagnostic performance of our proposed method across various abnormalities, we introduce two new validation datasets in addition to existing public benchmarks: CT-RATE-LT, consisting of 15 types of abnormalities, and RAD-ChestCT-LT, which encompasses 56 previously untested categories from the RAD-ChestCT dataset and naturally constitutes an open-set scenario.} As shown in Table~\ref{tab:comparison_combined}, we perform a comparative analysis of the performance of the open-source models and our proposed method on these two datasets. The results reveal that our method consistently outperforms other approaches in zero-shot diagnostic performance across a wide range of abnormalities. \textcolor{black}{This demonstrates that our framework generalizes well to unseen abnormalities under open-set conditions. We further validate on the INSPECT test set with contrast-enhanced CTPA cases for PE diagnosis in Table~\ref{tab:inspect_binary}, providing an out-of-distribution evaluation.} These improvements can be attributed to the effective vision-language alignment achieved by combining semantic summarization and CMKI module.

\begin{table}[t]
    \centering
    \renewcommand{\arraystretch}{1.}
    \scriptsize
    \caption{Volume-to-volume retrieval performance.}
    \begin{tabular}{@{}l lccc@{}}
    \toprule
    Dataset       & Method     & MAP@5   & MAP@10   & MAP@50 \\
    \midrule
    \multirow{6}{*}{CT-RATE} 
                  & CT-Net~\cite{draelos2021machine}$^{*}$     & 59.4$\pm$0.5    & 48.1$\pm$0.4     & 40.7$\pm$0.3   \\
                  & VocabFine~\cite{hamamci2024foundation}$^{*}$   & 68.3$\pm$0.4    & 57.2$\pm$0.4     & 48.8$\pm$0.3   \\
                  & ClassFine~\cite{hamamci2024foundation}$^{*}$   & 67.9$\pm$0.4    & 56.8$\pm$0.4     & 48.5$\pm$0.3   \\
                  & CT-CLIP~\cite{hamamci2024foundation}$^{*}$     & 68.3$\pm$0.4    & 57.2$\pm$0.4     & 48.9$\pm$0.3   \\
                  & Merlin~\cite{blankemeier2024merlin}$^*$     & 62.6$\pm$0.5    & 51.3$\pm$0.4     & 43.9$\pm$0.3   \\
                  & \algname   & \textbf{70.4$\pm$0.4} & \textbf{60.2$\pm$0.3} & \textbf{52.3$\pm$0.3} \\
    \midrule
    \multirow{6}{*}{RAD-ChestCT} 
                  & CT-Net~\cite{draelos2021machine}$^{*}$     & 71.2$\pm$0.3    & 58.9$\pm$0.3     & 52.4$\pm$0.2   \\
                  & VocabFine~\cite{hamamci2024foundation}$^{*}$   & 73.8$\pm$0.3    & 63.1$\pm$0.3     & 56.4$\pm$0.2   \\
                  & ClassFine~\cite{hamamci2024foundation}$^{*}$   & 73.8$\pm$0.3    & 63.1$\pm$0.3     & 56.3$\pm$0.2   \\
                  & CT-CLIP~\cite{hamamci2024foundation}$^{*}$     & 72.3$\pm$0.3    & 60.9$\pm$0.3     & 53.8$\pm$0.2   \\
                  & Merlin~\cite{blankemeier2024merlin}$^*$     & 74.0$\pm$0.3    & 63.3$\pm$0.3     & 56.6$\pm$0.2   \\
                  & \algname   & \textbf{75.7$\pm$0.3} & \textbf{65.7$\pm$0.2} & \textbf{59.1$\pm$0.2} \\
    \bottomrule
    \end{tabular}
    \label{tab:volume2volume}
    \vspace{-4mm}
\end{table}

\begin{table}[t]
\centering
\setlength{\tabcolsep}{4pt} 
\renewcommand{\arraystretch}{1.}
\scriptsize
\caption{\textcolor{black}{Cross-modal retrieval performance on CT-RATE.}}
\begin{tabular}{@{}lcccc@{}}
\toprule
Method  & Recall@5 & Recall@10 & Recall@50 & Recall@100 \\
\midrule
\textcolor{lightgray}{\textit{Report-to-Volume Retrieval}} \\
VocabFine~\cite{hamamci2024foundation}$^{*}$   & 0.2$\pm$0.1 & 0.5$\pm$0.1 & 2.2$\pm$0.3 & 4.4$\pm$0.4 \\
CT-CLIP~\cite{hamamci2024foundation}$^{*}$     & 2.9$\pm$0.3 & 5.2$\pm$0.4 & 17.5$\pm$0.7 & 28.3$\pm$0.8 \\
Merlin~\cite{blankemeier2024merlin}$^*$      & 1.5$\pm$0.2 & 2.7$\pm$0.3 & 7.7$\pm$0.5 & 12.7$\pm$0.6 \\
X2CT-CLIP~\cite{you2025x2ct}   & 4.8 & 7.7 & --   & --   \\
\algname    & \textbf{15.0$\pm$0.7} & \textbf{22.2$\pm$0.8} & \textbf{44.4$\pm$0.9} & \textbf{56.9$\pm$0.9} \\
\textcolor{lightgray}{\textit{Volume-to-Report Retrieval}} \\
VocabFine~\cite{hamamci2024foundation}$^{*}$   & 0.7$\pm$0.2 & 1.3$\pm$0.2 & 4.3$\pm$0.4 & 6.9$\pm$0.5 \\
CT-CLIP~\cite{hamamci2024foundation}$^{*}$     & 2.6$\pm$0.3 & 4.7$\pm$0.4 & 18.1$\pm$0.7 & 28.4$\pm$0.8 \\
Merlin~\cite{blankemeier2024merlin}$^*$      & 1.3$\pm$0.2 & 2.6$\pm$0.3 & 8.1$\pm$0.5 & 12.8$\pm$0.6 \\
\algname    & \textbf{18.5$\pm$0.7} & \textbf{25.5$\pm$0.8} & \textbf{49.0$\pm$0.9} & \textbf{61.0$\pm$0.9} \\
\bottomrule
\end{tabular}
\label{tab:crossmodal_retrieval}
\vspace{-4mm}
\end{table}

\subsection{Volume-to-Volume Retrieval Task}
\label{sec:VTV}
In the volume-to-volume retrieval task, cosine similarity is used to rank reference volumes by their proximity to a query volume in a shared latent space. The latent embeddings are generated by a vision encoder trained with a contrastive learning framework to align semantic features across volumes. Relevance between the query and retrieved volumes is measured based on the overlap of their abnormality labels. The experiments are conducted on both the internal and  external validation to ensure robustness under different distributions.

As shown in Table~\ref{tab:volume2volume}, our method achieves superior performance, which can be attributed to the cross-modal interaction module proposed in this study. By constructing the CMKB as a bridge for image-text features, our method effectively projects features into a shared space, significantly reducing the discrepancies between the image and text modalities. As a result, our method delivers outstanding performance in the volume-to-volume retrieval task.

\textcolor{black}{
\subsection{Cross-modal Retrieval Tasks}
\label{sec:RTV}
Cross-modal retrieval is conducted in two directions: \emph{report-to-volume} and \emph{volume-to-report}. Unlike the training phase where semantic summarization is introduced to facilitate alignment, here retrieval is conducted directly using the original reports. Retrieval is based on cosine similarity between text and image embeddings, where candidates are ranked and the top-$P$ volumes or reports are selected. Since RAD-ChestCT does not include radiology reports, all evaluations are carried out on the CT-RATE dataset.   
}

\textcolor{black}{
As shown in Table~\ref{tab:crossmodal_retrieval}, our method achieves the best performance across all metrics, substantially surpassing CT-CLIP, the strongest baseline. For report-to-volume retrieval, Recall@10 improves from 5.2 to 22.2 and Recall@100 from 28.3 to 56.9; for volume-to-report retrieval, Recall@10 rises from 4.7 to 25.5 and Recall@100 from 28.4 to 61.0. These gains stem from the proposed CMKB, which bridges image and text features in a shared space, reducing modality gaps and improving retrieval alignment.
}

\subsection{Ablation Study}
To evaluate the contributions of each module to the performance of \algname, we conduct ablation experiments on the zero-shot abnormality diagnosis and the report-to-volume retrieval tasks. The results are presented in Table~\ref{tab:ablation_study} and Table~\ref{tab:ablation_report2volume}, with CLIP used as the baseline. The proposed method incorporates two core modules: semantic summarization and CMKI. 

First, the independent introduction of the semantic summarization module leads to a significant performance boost (AUC improves from 65.0 to 67.9 in Table~\ref{tab:ablation_study}, and Recall@50 increases from 13.1 to 18.4 in Table~\ref{tab:ablation_report2volume}). By leveraging an LLM to perform semantic summarization of the reports, this module effectively reduces the complexity of textual feature learning. Consequently, it improves image-text alignment quality, leading to improved zero-shot diagnostic and retrieval performance.

Second, the independent introduction of the CMKI module similarly results in a substantial performance enhancement compared to the baseline (AUC is improved to 68.0 from 65.0 in Table~\ref{tab:ablation_study}, and Recall@50 is improved to 21.8 from 13.1 in Table~\ref{tab:ablation_report2volume}). Specifically, in the report-to-volume retrieval task, the results demonstrate notable improvements, which can be attributed to the introduction of the CMKB as a bridge between image and text features. The CMKI module facilitates implicit alignment for unpaired image-text features by leveraging the shared latent space constructed by the CMKB. This process enables cross-modal information interaction, reduces discrepancies between modalities, and preserves the unique characteristics of each modality, thereby improving the alignment of image-text features and enhancing retrieval performance. In the zero-shot diagnostic task, the CMKI module effectively reduces the modality gap in vision-language alignment, enabling precise abnormality classification and demonstrating its capability to enhance cross-modal alignment for robust diagnostic inference.

Finally, by combining the semantic summarization and CMKI modules, \algname achieves the best performance (AUC: 70.0 in Table~\ref{tab:ablation_study}, Recall@50: 28.6 in Table~\ref{tab:ablation_report2volume}). The semantic summarization module simplifies textual learning and extracts high-level semantic information, while the CMKI module bridges the modality gap through implicit alignment and cross-modal interaction. Together, these modules enhance vision-language alignment, leading to improved performance in both zero-shot diagnosis and report-to-volume retrieval tasks.

    \begin{table}[htbp]
    \centering
    \setlength{\tabcolsep}{5pt}
    \renewcommand{\arraystretch}{1.}
    \scriptsize
    \caption{Ablation study on the Semantic Summarization (SS) and the CMKI module for zero-shot abnormality diagnosis on the RAD-ChestCT dataset.}
   \begin{tabular}{@{}c c ccccccc@{}}
    \toprule
    SS & CMKI & AUC  & ACC & F1 & Prec & mAP & Rec@1 & Prec@3  \\
    \midrule
    ~ & ~ & 65.0  & 61.1 & 65.4 & 35.7 &  39.3 & 12.9 & 38.9  \\
    \checkmark & ~ & 67.9  & 64.0 & 67.9 & 37.2 &  40.9 & 13.3 & 42.2  \\
    ~ & \checkmark & 68.0  & 64.0 & 67.9 & 38.0 &  41.7 & 12.3 & 40.2 \\
    \checkmark & \checkmark & \textbf{70.0}  & \textbf{65.5} & \textbf{69.3} & \textbf{39.1} &  \textbf{43.0} &  \textbf{14.3} & \textbf{44.0} \\
    \bottomrule
    \end{tabular}
    \label{tab:ablation_study}
    \vspace{-4mm}
    \end{table}

\begin{table}[htbp]
    \centering
    \renewcommand{\arraystretch}{1.}
    \scriptsize
    \caption{Ablation study on the impact of the Semantic Summarization (SS) and the CMKI module for the report-to-volume retrieval task on the CT-RATE validation dataset.}
   \begin{tabular}{@{}c c cccc@{}}
    \toprule
    SS & CMKI & Recall@5 &  Recall@10 &  Recall@50 &  Recall@100  \\
    \midrule
    ~ & ~ & 2.2  & 3.7 & 13.1 & 22.2 \\
    \checkmark & ~ & 2.2 & 4.1 &  18.4 & 29.1  \\
    ~ & \checkmark & 3.4  & 5.8 & 21.8 & 34.6  \\
    \checkmark & \checkmark & \textbf{5.8} & \textbf{10.1} & \textbf{28.6} & \textbf{42.0} \\
    \bottomrule
    \end{tabular}
    \label{tab:ablation_report2volume}
    \vspace{-4mm}
\end{table}

\subsection{Robustness of Semantic Summarization}
\textcolor{black}{To evaluate the reliability of semantic summarization in our framework, we conduct two experiments to assess the reliability of semantic summarization in our framework.}  

\textcolor{black}{
\textbf{(a) Summarizer comparison.}  
We compare summarizers of different scales as well as domains, including general-domain models (e.g., Qwen-2.5, Llama-3, GPT-4, DeepSeek) and medical-domain models (e.g., Lingshu, Medgemma).
}

\textcolor{black}{
\textbf{(b) Report perturbations.}  
We introduce three types of noise before summarization: (1) style conversion with shortened and lengthened expressions, (2) grammatical errors in five sentences per report, and (3) random removal of 30\% or 40\% of sentences. The perturbed reports are then summarized using semantic summarization module based on GPT-4 for downstream alignment and zero-shot classification.
}

\textcolor{black}{
Results in Table~\ref{tab:summ_robust_clip_metrics} reveal three consistent trends:  
(1) models of different scales show comparable summarization quality, and larger models do not consistently outperform smaller ones;  
(2) medical-domain models provide performance on par with general-domain counterparts, indicating limited domain-specific advantage; and  
(3) the pipeline remains robust under report perturbations, maintaining competitive accuracy even with grammatical errors or missing sentences.  
These findings demonstrate that our approach is robust and generalizable, even in noisy or imperfect clinical reporting scenarios.
}

\begin{table}[htbp]
\centering
\setlength{\tabcolsep}{3pt}
\scriptsize
\caption{\textcolor{black}{Zero-shot diagnostic performance under different summarizers (a) and under report perturbations (b).}}
\begin{tabular}{@{}lccccccc@{}}
\toprule
\multicolumn{8}{c}{\textbf{(a) Summarizer Comparison on Original Reports}} \\
\midrule
Summarizer & AUC & ACC & F1 & Prec & mAP & Rec@1 & Prec@3 \\
\midrule
\textcolor{lightgray}{\textit{Smaller-scale Language Models}} \\

Qwen-2.5-7B~\cite{team2024qwen2}          & 69.1 & 64.6 & 68.5 & 38.1 & 41.8 & \textbf{15.3} & \textbf{45.3} \\
Lingshu-7B~\cite{xu2025lingshu}          & 69.2 & 64.9 & 68.7 & 38.3 & 42.4 & 9.9 & 43.0 \\
Llama-3-8B~\cite{llama3modelcard}        & 68.6 & 65.1 & 68.9 & 38.4 & 42.0 & 12.3 & 44.3 \\
Medgemma-4b~\cite{sellergren2025medgemma}       & 68.6 & 64.7 & 68.5 & 37.8 & 41.7 & 10.8 & 43.7 \\

\textcolor{lightgray}{\textit{Larger-scale Language Models}} \\

Medgemma-27b~\cite{sellergren2025medgemma}       & 69.2 & 64.8 & 68.6 & 38.1 & 42.5 & 14.8 & 43.2 \\
Deepseek-V3~\cite{liu2024deepseek}       & 68.6 & 65.9 & 69.5 & 38.5 & 42.6 & 13.5 & 42.9 \\
GPT-4 Turbo & \textbf{70.0}  & \textbf{65.5} & \textbf{69.3} & \textbf{39.1} &  \textbf{43.0} &  14.3 & 44.0 \\
\midrule
\multicolumn{8}{c}{\textbf{(b) Robustness Under Report Perturbations}} \\
\midrule
Perturbation & AUC & ACC & F1 & Prec & mAP & Rec@1 & Prec@3 \\
\midrule
Baseline  & \textbf{70.0}  & 65.5 & 69.3 & \textbf{39.1} &  43.0 &  14.3 & 44.0 \\
Shortened expressions      & 69.9 & \textbf{65.8} & \textbf{69.5} & 38.9 & \textbf{43.3} & 14.0 & 43.7 \\
Lengthened expressions      & 68.3 & 64.6 & 68.5 & 37.8 & 41.9 & \textbf{15.4} & \textbf{45.8} \\
Grammatical errors      & 67.6 & 63.6 & 67.5 & 37.5 & 41.1 & 13.5 & 43.0 \\
Missing sentences (30\%)    & 67.4 & 64.5 & 68.3 & 37.6 & 41.1 & 10.3 & 40.4 \\
Missing sentences (40\%)    & 66.9 & 64.4 & 68.2 & 37.7 & 41.4 & 9.6 & 39.2 \\
\bottomrule
\end{tabular}
\label{tab:summ_robust_clip_metrics}
\vspace{-4mm}
\end{table}

\begin{figure*}[htbp]
  \centering
    \includegraphics[width=1.0\linewidth]{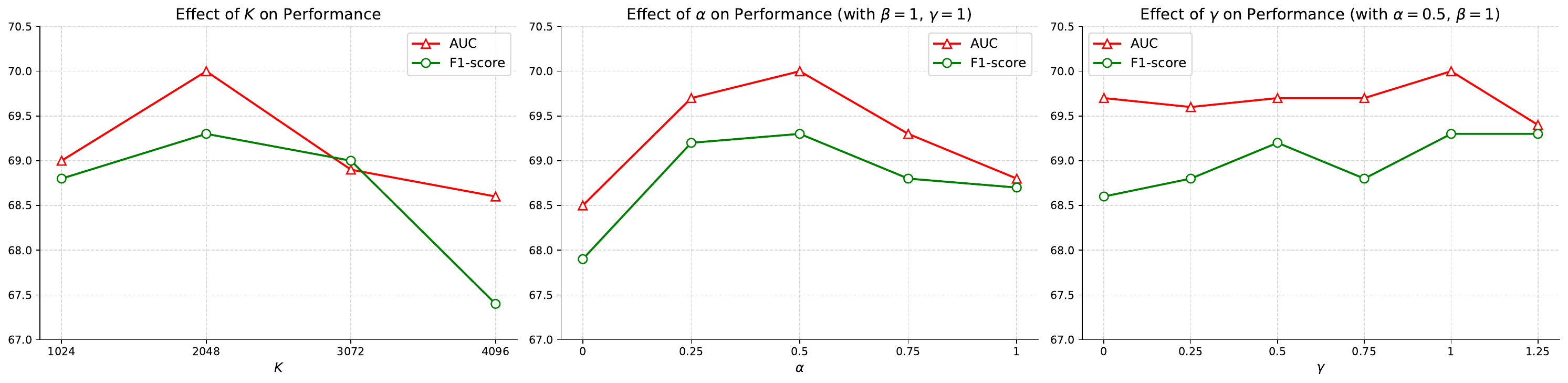}
    \caption{\textcolor{black}{Ablation results on the RAD-ChestCT dataset, showing the effect of varying $K$ in CMKB as well as the impact of the hyperparameters $\alpha$ and $\gamma$ in the total alignment loss.}}

    \label{fig:pkp}
\end{figure*}

    \begin{table}[htbp]
    \centering
    \setlength{\tabcolsep}{3pt}
    \renewcommand{\arraystretch}{1.}
    \scriptsize
    \caption{\textcolor{black}{Ablation study of CMKB initialization strategies, including random distributions and pretrained embeddings, evaluated on RAD-ChestCT for zero-shot diagnosis.}}
   \begin{tabular}{@{}l  ccccccc@{}}
    \toprule
    CMKB Initialization  & AUC  & ACC & F1 & Prec & mAP & Rec@1 & Prec@3  \\
    \midrule
    \textcolor{lightgray}{\textit{Random initialization}} \\
    Gaussian Random &  68.7  & 64.6 & 68.4 & 38.3 &  41.6 & 13.8 & 43.4  \\
    Xavier Random&  68.8  & 64.7 & 68.6 & 38.2 &  42.0 & \textbf{14.8} & 43.0  \\
    Uniform Random &  68.5  & 64.2 & 68.1 & 37.6 &  41.6 & 13.7 & 43.1 \\
    Orthogonal Random &  \textbf{70.0}  & \textbf{65.5} & \textbf{69.3} & \textbf{39.1} &  \textbf{43.0} &  14.3 & \textbf{44.0} \\
    \textcolor{lightgray}{\textit{Embedding-based initialization}} \\
    Image Embedding & 68.4  & 64.4 & 68.3 & 38.3 &  42.4 & 10.6 & 43.7 \\
    Text Embedding & 63.9  & 61.7 & 65.8 & 35.4 &  37.5 & 11.7 & 40.4 \\
    Cross Embedding & 65.4  & 61.9 & 66.0 & 36.0 &  39.1 & 14.1 & 42.0 \\
    UMLS~\cite{UMLS2024AA} & 68.7  & 64.5 & 68.4 & 37.9 &  41.8 & 14.1 & 42.2 \\
    RadLex~\cite{langlotz2006radlex} & 69.4  & 65.5 & 69.2 & 38.7 &  43.3 & 14.0 & 43.1 \\
    
    \bottomrule
    \end{tabular}
    \label{tab:pretrain_dictionary}
    \vspace{-4mm}
    \end{table}

\subsection{Effect of Initialization Strategies in CMKB}
\textcolor{black}{
We evaluate the effect of different initialization schemes for CMKB, comparing random distributions and embedding-based }\textcolor{black}{priors. Random strategies include Gaussian, Xavier, Uniform, and Orthogonal initialization. For embedding-based strategies, we use features extracted from the vision and text encoders pretrained with M3AE. \emph{Image} initialization applies $k$-means clustering on vision encoder features from the training set, while \emph{Text} uses text encoder features. \emph{Cross} combines both modalities before clustering to form cross-modal centroids. In addition, \emph{UMLS}~\cite{UMLS2024AA} and \emph{RadLex}~\cite{langlotz2006radlex} are treated as textual corpora, encoded with the text encoder and clustered into pretrained embeddings. As shown in Table~\ref{tab:pretrain_dictionary}, orthogonal random initialization yields the best results among random schemes, whereas RadLex provides the best embedding-based initialization, demonstrating the benefit of domain-specific ontologies for CMKB. The relatively poor performance of \emph{Text} initialization may be due to the repetitive and template-driven nature of training reports, which could lead to clustered embeddings lacking diversity.
}

\subsection{Hyperparameter Sensitivity Analysis}
\textbf{Effect of $K$.}  
The hyper-parameter \(K\) in CMKB, which defines the number of basis vectors, plays a pivotal role in determining its spatial representational capacity. To evaluate the influence of \(K\), we conduct experiments with values $K$ = \{1024, 2048, 3072, 4096\}. As illustrated in Fig.~\ref{fig:pkp}, increasing \(K\) enhances the spatial coverage of CMKB, enabling the model to capture a broader range of features and improve image-text alignment quality. However, this improvement diminishes and eventually saturates as \(K\) becomes excessively large. In such scenarios, an overabundance of basis vectors can hinder cross-modal information interaction. For instance, when \(K\) is excessively large, the model may reconstruct features independently within each modality, effectively bypassing the role of CMKI and nullifying cross-modal interaction. Based on these observations, we set \(K = 2048\) for all subsequent experiments, as it strikes an optimal balance between alignment performance and computational efficiency.

\textcolor{black}{\textbf{Effect of loss weights.} We further study the loss hyper-parameters \(\alpha\) and \(\gamma\), with \(\beta\) fixed to 1. As illustrated in Fig.~\ref{fig:pkp}, moderate values of \(\alpha\) (around 0.5) enhance performance by} \textcolor{black}{strengthening reconstruction constraints, while larger values lead to performance degradation. Setting \(\alpha=0\) markedly reduces both AUC and F1, confirming that the reconstruction loss \(\ell_{\text{MSE}}\) is essential. Similarly, \(\gamma\) controls the contribution of }\textcolor{black}{the InfoNCE on reconstructed features \(\ell_{\text{INFO-R}}\). Small to moderate values improve alignment, with the best results obtained near \(\gamma=1\). Removing this term (\(\gamma=0\)) results in weaker alignment and lower performance, demonstrating its necessity. For completeness, we also examined the case of removing the primary contrastive alignment loss by setting \(\beta=0\) (i.e., ablating \(\ell_{\text{INFO}}\)). This caused a drastic performance drop (AUC 51.8, F1 54.3), far below all other variants, clearly indicating that \(\ell_{\text{INFO}}\) is indispensable for maintaining alignment quality.}

\begin{figure}
  \centering
    \includegraphics[width=0.8\linewidth]{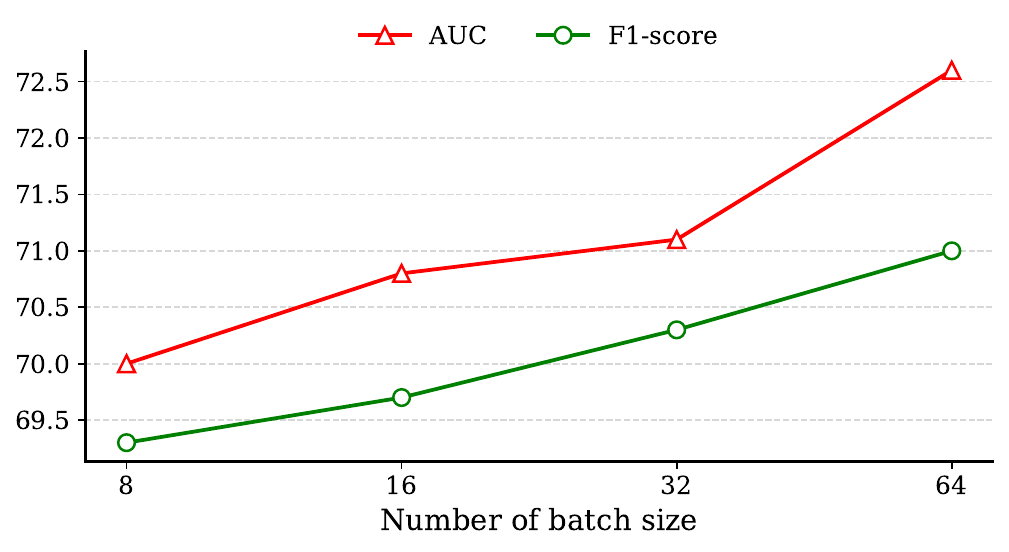}
    \caption{\textcolor{black}{Results of varying batch size for BrgSA on RAD-ChestCT dataset.}} 
    \label{fig:bs}
    \vspace{-4mm}
\end{figure}

\textcolor{black}{\textbf{Effect of batch size.}  
We also examine the influence of batch size on InfoNCE optimization, as shown in Fig.~\ref{fig:bs}. Larger batch sizes consistently yield better performance: for example, the AUC improves from 70.0 at batch size 8 to 72.6 at batch size 64, while the F1 score rises from 69.3 to 71.0. This aligns with the intuition from CLIP training that larger batches provide more diverse negative samples within each iteration, strengthening contrastive alignment between modalities. Although hardware resources limit the maximum batch size in our experiments, future work could explore scaling to larger batches for further gains.}

\begin{figure*}[htbp]
  \centering
    \includegraphics[width=1.0\linewidth]{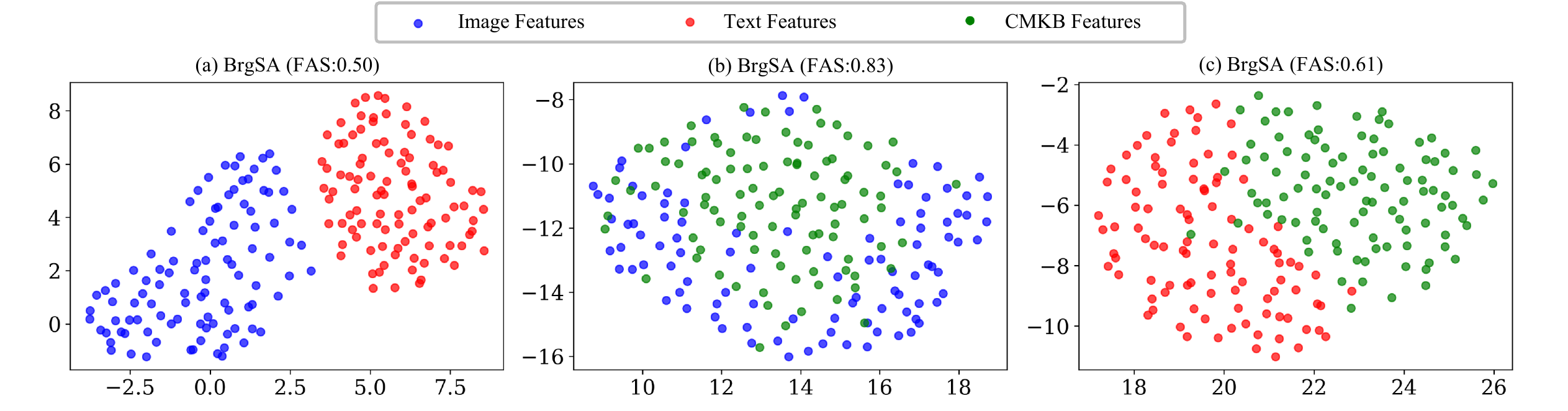}
    \caption{UMAP visualization of features. Cosine similarity is used to evaluate the alignment between image and text features. The text features are generated using generic descriptive texts to ensure that all images can be matched to all texts. (a)--(c) represent features trained by \algname.} 
    \label{fig:umap}
    \vspace{-4mm}
\end{figure*}

\begin{table}[htbp]
\centering
\setlength{\tabcolsep}{3.5pt} 
\renewcommand{\arraystretch}{1.}
\scriptsize
\caption{\textcolor{black}{Ablation on vision and text encoders for CLIP-based zero-shot diagnosis. 
We control one encoder while varying the other to ensure fair single-variable comparison.}}
\begin{tabular}{@{}lccccccc@{}}
\toprule
\multicolumn{8}{c}{\textbf{(a) Vision Encoder Exploration (Fixed Text: PubMedBERT)}} \\
\midrule
Vision Encoder & AUC & ACC & F1 & Prec & mAP & Rec@1 & Prec@3 \\
\midrule
ResNet18~\cite{he2016deep}       & 70.8 & 66.6 & 70.2 & 39.7 & 45.2 & 14.6 & 46.2 \\
ResNet34~\cite{he2016deep}       & 70.4 & 66.3 & 69.9 & 39.7 & 43.6 & 13.0 & 45.6 \\
ResNet50~\cite{he2016deep}       & 70.4 & 66.0 & 69.8 & 39.4 & 43.6 & 14.7 & 46.8 \\
SwinVit-B~\cite{liu2021swin}         & 72.2 & \textbf{67.7} & \textbf{71.1} & 40.5 & 47.3 & 13.8 & 47.1 \\
ViT-B~\cite{dosovitskiy2020image}       & \textbf{72.6}  & \textbf{67.7} & \textbf{71.1} & \textbf{40.9} &  \textbf{47.6} &  \textbf{14.9} & \textbf{46.8} \\
\midrule
\multicolumn{8}{c}{\textbf{(b) Text Encoder Exploration (Fixed Vision: ViT-B/16)}} \\
\midrule
Text Encoder     & AUC & ACC & F1 & Prec & mAP & Rec@1 & Prec@3 \\
\midrule

BioClinicalBERT~\cite{alsentzer-etal-2019-publicly}  & 72.7 & 68.0 & 71.4 & 41.6 & 47.5 & 13.8 & 46.0 \\
CXRBERT~\cite{boecking2022making}    & \textbf{74.2} & \textbf{68.6} & \textbf{72.0} & \textbf{42.2} & \textbf{49.4} & \textbf{14.9} & \textbf{48.7} \\
BioGPT~\cite{luo2022biogpt}         & 64.9 & 62.8 & 66.9 & 36.2 & 38.7 & 13.0 & 42.2 \\
PubMedBERT~\cite{gu2021domain}       & 72.6  & 67.7 & 71.1 & 40.9 &  47.6 &  14.3 & 46.8 \\   
\bottomrule
\end{tabular}
\label{tab:encoder_ablation}
\vspace{-4mm}
\end{table}

\subsection{Exploration of Vision and Text Encoders}  
\textcolor{black}{
Table~\ref{tab:encoder_ablation} shows the impact of different vision and text backbones. 
For vision encoders, ViT-B performs best, benefiting from global self-attention that captures long-range dependencies in CT volumes. 
For text encoders, while we initially used PubMedBERT~\cite{gu2021domain}, CXRBERT~\cite{boecking2022making} achieves the best results due to its pretraining on radiology reports. 
BioClinicalBERT is competitive but less radiology-specific, and BioGPT underperforms as a generative model. 
Overall, }\textcolor{black}{ViT-B and CXRBERT prove most effective, and we adopt them as the final encoders in our framework.
}

\subsection{Inference Efficiency Benchmark}
\textcolor{black}{We benchmark diagnostic performance and efficiency on RAD-ChestCT (Table~\ref{tab:deploy_benchmark}). Our method achieves the highest AUC (74.2) with comparable latency (43.1 ms/study) and memory usage (1.7 GB), clearly outperforming prior approaches in accuracy. While fVLM~\cite{shui2025large} shows slightly lower latency and GPU memory, its pipeline requires an additional automatic segmentation step whose computational cost is not included in the table, making its reported efficiency less favorable in practice. Furthermore, by directly running inference with reduced input resolutions, our model maintains competitive performance (AUC 67.7–72.2) while reducing latency to 21.7 ms and memory to 1.1 GB, highlighting its strong adaptability for real-world deployment.}

\begin{table}[htbp]
\centering
\setlength{\tabcolsep}{1pt} 
\renewcommand{\arraystretch}{1.}
\scriptsize
\caption{\textcolor{black}{Evaluation on RAD-ChestCT: AUC with inference efficiency (time per study and peak GPU memory).}}

\begin{tabular}{@{}lcccccc@{}}
\toprule
Method & Height & Width & Slices & AUC & Time (ms/study)~$\downarrow$ & GPU Mem (GB)~$\downarrow$ \\
\midrule

CT-CLIP~\cite{hamamci2024foundation}      & 480 & 480 & 240 & 62.9$\pm$1.3 & 96.9$\pm$1.9 & 4.1 \\
Merlin~\cite{blankemeier2024merlin}       & 224 & 224 & 160 & 64.4$\pm$1.2 & 176.3$\pm$0.2 & 2.1 \\
fVLM~\cite{shui2025large}         & 352 & 288 & 112 & 68.0$\pm$1.2 & 42.2$\pm$0.7 & 1.6 \\
\algname     & 224 & 224 & 112 & \textbf{74.2$\pm$1.0} & 43.1$\pm$0.5 & 1.7 \\
\algname     & 224 & 224 & 64  & 72.2$\pm$1.1 & 24.1$\pm$0.7 & 1.4 \\
\algname     & 112 & 112 & 112 & 69.6$\pm$1.1 & 22.6$\pm$1.0 & 1.2 \\
\algname     & 112 & 112 & 64  & 67.7$\pm$1.2 & \textbf{21.7$\pm$0.3} & \textbf{1.1} \\

\bottomrule
\end{tabular}
\label{tab:deploy_benchmark}
\vspace{-4mm}
\end{table}

\begin{figure}[htbp]
  \centering
    \includegraphics[width=0.9\linewidth]{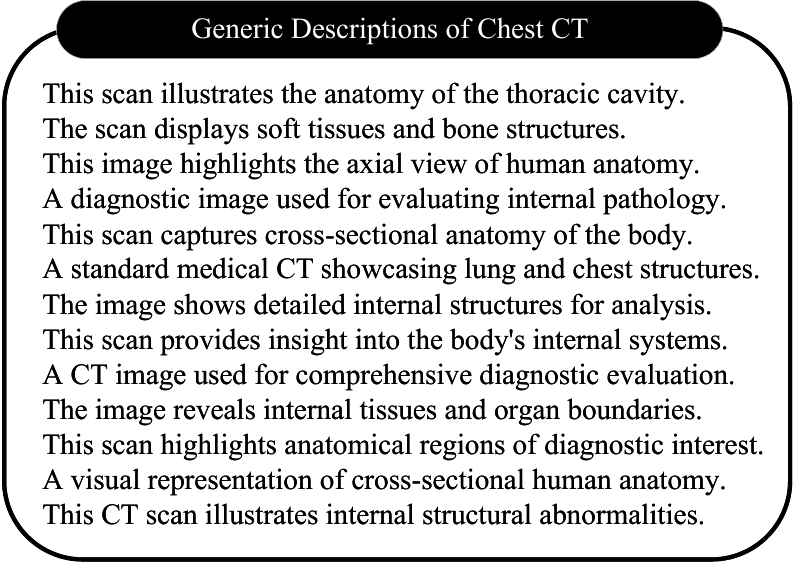}
    \caption{Generic descriptions of chest CT generated by a LLM for visualizing text features.} 
    \label{fig:gd}
    \vspace{-4mm}
\end{figure}

\subsection{Visualization of Features}
Fig.~\ref{fig:motivation} and Fig.~\ref{fig:umap} illustrate the UMAP visualization of image, text, and CMKB features. To assess the effectiveness of vision-language alignment, we ensure that all text descriptions are aligned with all CT volumes. As shown in Fig.~\ref{fig:gd}, we leverage GPT-4 to generate unbiased, generic descriptions of chest CT volumes as text samples for feature extraction. During the UMAP computation, cosine similarity is employed to measure the distances between different features, representing the degree of alignment among them. 
\textcolor{black}{To quantify this alignment, we first introduce the silhouette score~\cite{rousseeuw1987silhouettes}, originally used to evaluate cluster separation. For each sample $i$, let $a(i)$ denote the mean intra-cluster distance and $b(i)$ the mean nearest-cluster distance. The silhouette score is then defined as}
\begin{equation}
\textcolor{black}{s(i) = \frac{b(i) - a(i)}{\max\{a(i),\, b(i)\}}, \quad S = \frac{1}{N}\sum_{i=1}^N s(i).}
\end{equation}
\textcolor{black}{In our setting, the three feature sets (image, text, and CMKB) are predefined rather than obtained by unsupervised clustering. Hence, each sample has a fixed cluster assignment and no mis-clustering can occur. Consequently, the intra-cluster distance $a(i)$ is never larger than the inter-cluster distance $b(i)$, ensuring that $s(i)$ remains non-negative and $S$ lies in $[0,1]$.}
\textcolor{black}{To adapt this metric for evaluating vision-language alignment, we define the Feature Alignment Score (FAS) as}
\begin{equation}
\textcolor{black}{\text{FAS} = 1 - S,}
\label{eq:fas}
\end{equation}
\textcolor{black}{where a score closer to 1 indicates better alignment, while a score closer to 0 indicates poorer alignment.}

Comparing Fig.~\ref{fig:motivation}~(b) and Fig.~\ref{fig:umap}~(a), we observe that \algname significantly reduces the gap between image and text feature spaces, resulting in higher FAS values (0.37 vs. 0.50). This demonstrates that CMKI effectively enhances vision-language alignment. Further, Fig.~\ref{fig:umap}~(b) and Fig.~\ref{fig:umap}~(c) reveal that CMKB features align more closely with image features than with text features. This discrepancy can be attributed to the high-level abstraction and relatively uniform nature of text features, whereas image features are richer and more diverse. The greater variability in image feature representation makes CMKB features gravitate toward image features. Finally, as shown in Fig.~\ref{fig:motivation}~(c), CMKB features act as a bridge between image and text features, connecting the two feature spaces and facilitating their interaction. This bridging role effectively improves the overall vision-language alignment, further validating the importance of CMKB in this process.

\begin{figure}[t]
  \centering
    \includegraphics[width=0.95\linewidth]{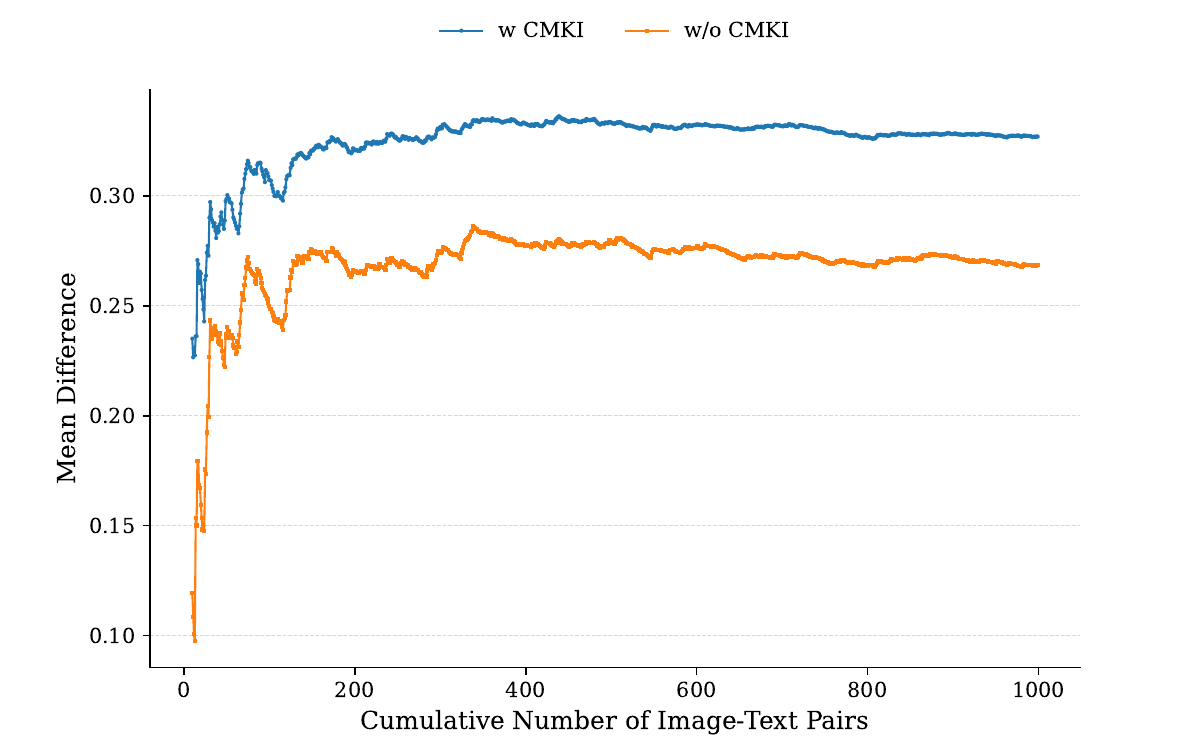}
    \caption{Comparison of mean differences between paired and unpaired image-text samples on CT-RATE validation dataset.} 
    \label{fig:differ}
    \vspace{-4mm}
\end{figure}

\subsection{Mean Difference for Pair-unpair Discrimination}

The goal of CLIP is to increase the similarity of paired image-text samples while decreasing the similarity of unpaired image-text samples. To evaluate the quality of image-text alignment, we define the mean difference as follows:
\begin{equation}
\Delta_{\text{mean}} = \frac{1}{n(n-1)} \sum_{i=1}^n \sum_{j \neq i} \left( M(i, i) - M(i, j) \right),
\end{equation}
where \( M(i, i) \) represents the similarity of the \( i \)-th paired image-text sample, \( M(i, j) \) represents the similarity between the \( i \)-th image and the \( j \)-th text, and \( n \) is the total number of image-text pairs. The results are evaluated on the CT-RATE validation set to demonstrate the effectiveness of the alignment.

Fig.~\ref{fig:differ} shows the variation of mean difference between paired and unpaired image-text samples. The horizontal axis represents the cumulative number of image-text pairs, and the vertical axis represents mean difference, \( \Delta_{\text{mean}} \). It can be observed that the methods incorporating the CMKI module achieve a higher \( \Delta_{\text{mean}} \) curve compared to those without the CMKI module. This indicates that the CMKI module leads to a greater similarity difference between paired and unpaired samples, enabling better discrimination between them. This improvement is attributed to the role of CMKB as a bridge between image and text features. By projecting features from both modalities into a shared latent space, the CMKB effectively reduces the discrepancy between modalities while preserving their unique characteristics. This enhances the vision-language alignment and improves the ability to differentiate paired and unpaired image-text samples.

\begin{figure}[t]
  \centering
    \includegraphics[width=0.95\linewidth]{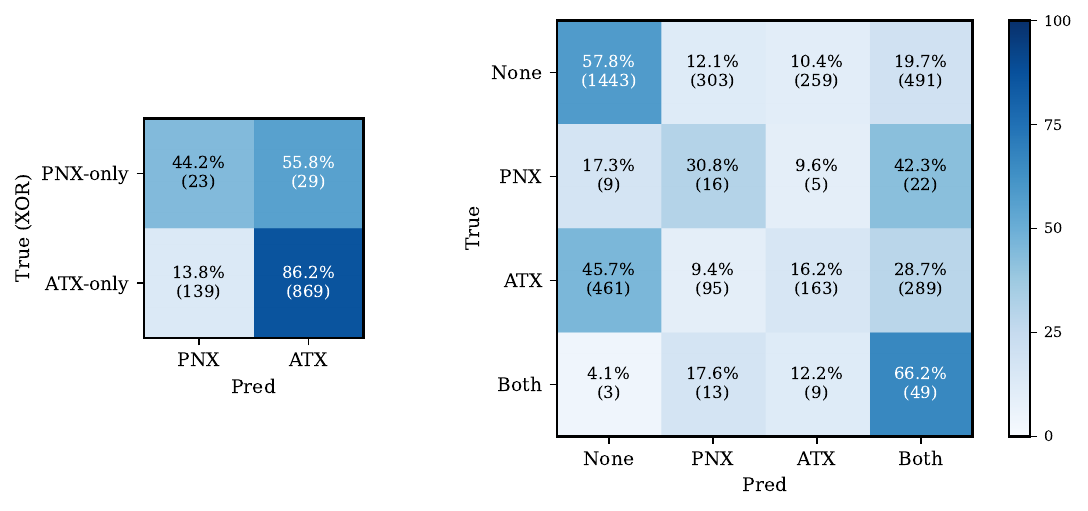}
    \caption{\textcolor{black}{Pairwise confusion analysis of Pneumothorax (PNX) and Atelectasis (ATX): 2$\times$2 XOR-subset (left) and 4$\times$4 joint states (right). Cells show row-normalized percentages with counts in parentheses.}}

    \label{fig:differ_confusion}
    \vspace{-4mm}
\end{figure}

\section{Discussion}

In this work, we present \algname, a bridged semantic alignment framework that closes the gap between 3D medical images and diagnostic texts for zero-shot disease recognition. \algname combines InfoNCE contrastive learning with CMKI-based masked cross-modal reconstruction to yield robust, interpretable alignment. We then discuss its clinical and technical impact, innovations, and limitations.

\subsubsection{Clinical Impact} Zero-shot learning offers a promising solution for diagnosing unseen diseases, particularly rare conditions that have very limited annotated samples and are difficult to handle using conventional supervised methods. Our proposed method, \algname, enhances vision-language alignment in 3D medical data and significantly improves zero-shot diagnostic performance. This improvement is of great clinical relevance for rare disease diagnosis, where prior examples are scarce. To support clinical validation, we construct CT-RATE-LT, a benchmark dataset featuring underrepresented abnormalities. Our model achieves strong results on both CT-RATE-LT and the RadChest-LT benchmark, demonstrating its robustness across different datasets. Moreover, unlike many prior methods focusing on 2D images, our approach is designed for 3D CT, a clinically critical imaging modality, further underscoring its potential for real-world medical applications. \textcolor{black}{We also provide confusion analyses of clinically related findings, such as pneumothorax versus atelectasis (Fig.~\ref{fig:differ_confusion}). The results indicate that while the model can distinguish the two abnormalities in many cases, confusion remains—particularly with pneumothorax being misclassified as atelectasis and atelectasis often under-detected. These findings highlight both the potential and the limitations for clinical application.}

\subsubsection{Technical Impact} We propose a \textit{simple yet effective} framework to enhance vision-language alignment in 3D medical data. At the core of our method propose the CMKB, which acts as a semantic bridge between image and text features. By enabling implicit interaction across modalities, CMKB reinforces shared semantics and alleviates modality-specific biases. This design leads to more accurate and robust feature alignment, which is critical for improving zero-shot diagnostic performance in 3D medical imaging tasks.

\subsubsection{Innovation} This work introduces two key innovations to advance zero-shot 3D medical image diagnosis. First, we propose the CMKI module, which implicitly aligns 3D visual and textual features by bridging the modality gap. This component complements contrastive learning and enables more stable and semantically coherent alignment across modalities. Second, we construct CT-RATE-LT, a benchmark test set comprising underrepresented abnormalities, to evaluate zero-shot performance on rare and low-frequency conditions.

\subsubsection{Limitations and Future Work} 
\begin{figure}[t]
  \centering
    \includegraphics[width=0.95\linewidth]{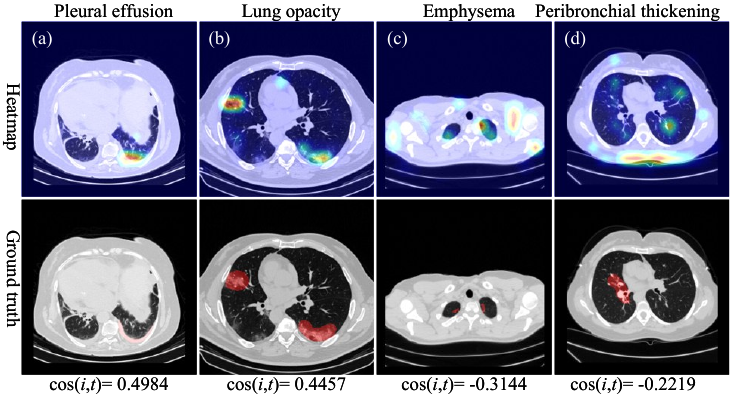}
    \caption{\textcolor{black}{Representative cases with attention heatmaps (top) and ground-truth masks (bottom). (a),(b) show successful predictions with high similarity scores for positive findings, while (c),(d) illustrate missed cases with low similarity despite positive ground truth.}}
    \label{fig:fail_cases}
    \vspace{-4mm}
\end{figure}

While our method demonstrates strong performance in zero-shot 3D medical diagnosis, several limitations remain. 
First, although the framework effectively integrates semantic summarization and cross-modal knowledge interaction for global alignment, it may still underrepresent fine-grained local features that are essential for detecting subtle or complex abnormalities in 3D volumes. 
Second, by contrasting success and failure cases, we find that the model occasionally misaligned its attention, overlooking true pathological regions and resulting in missed detections in low-confidence predictions (Fig.~\ref{fig:fail_cases}). 
Third, our current focus on zero-shot generalization leaves open the question of how well the method performs under limited supervision. 
\textcolor{black}{Fourth, since the current evaluation is limited to thoracic CT datasets, the method’s generalizability to other anatomical domains remains to be validated. 
Fifth, as our framework partially relies on LLM-based semantic summarization, potential biases inherent in pretrained LLMs may influence the generated representations. 
Sixth, the framework’s relatively high computational cost could pose deployment challenges in resource-constrained clinical environments. To address these limitations, future work will enhance local feature alignment and attention calibration to improve sensitivity to subtle abnormalities. 
We also plan to investigate adaptive learning strategies for broader generalization, while exploring lightweight architectures and bias-mitigation techniques to enable fair and efficient deployment in real-world clinical settings.
} 

\section{Conclusion}

This paper proposes a \algname framework to narrow the modality gap in 3D medical image-text alignment for zero-shot abnormality diagnosis. The \algname framework consists of a semantic summarization module and a CMKI module. The semantic summarization module leverages a large language model to summarize radiology reports, reducing the complexity of textual feature learning and generating high-level semantic representations. The CMKI module facilitates cross-modal interaction between image and text features, effectively reducing modality discrepancies while preserving their unique characteristics. Leveraging these components, the \algname framework effectively reduces the modality gap and achieves robust vision-language alignment. Experimental results validate its efficacy, demonstrating competitive performance in both zero-shot diagnosis and retrieval tasks across internal and external validation datasets.

\section*{References}
\bibliographystyle{IEEEtran}
\bibliography{main}

\renewcommand\thefigure{S\arabic{figure}} 
\setcounter{figure}{0} 
\renewcommand\thetable{S\arabic{table}} 
\setcounter{table}{0}

\end{document}